\documentclass[lettersize,journal]{IEEEtran}
\usepackage{amsmath,amsfonts}
\usepackage{algorithmic}
\usepackage{algorithm}
\usepackage{array}
\usepackage[caption=false,font=normalsize,labelfont=sf,textfont=sf]{subfig}
\usepackage{textcomp}
\usepackage{stfloats}
\usepackage{url}
\usepackage{verbatim}
\usepackage{graphicx}
\usepackage{cite}
\usepackage{hyperref}
\usepackage{pifont}
\usepackage{hyperref}
\usepackage{booktabs}
\hypersetup{hypertex=true,
colorlinks=true,
linkcolor=blue,
anchorcolor=blue,
citecolor=green}

\usepackage{hyperref}
\usepackage{booktabs}

\usepackage{graphicx} 
\usepackage{multirow} 
\usepackage{booktabs} 

\usepackage{tabularx} 
\usepackage{caption}
\usepackage{graphicx}
\usepackage{float} 
\usepackage{subcaption}

\begin{document}

\title{Self-Supervised State Space Model for Real-Time Traffic Accident Prediction Using eKAN Networks}

\author{Xin Tan, Meng Zhao~\IEEEmembership{}
\thanks{This paper was produced by the IEEE Publication Technology Group. They are in Piscataway, NJ.}
\thanks{Manuscript received April 19, 2021; revised August 16, 2021.}}

\markboth{Journal of \LaTeX\ Class Files,~Vol.~14, No.~8, August~2021}%
{Shell \MakeLowercase{\textit{et al.}}: A Sample Article Using IEEEtran.cls for IEEE Journals}


\maketitle

\begin{abstract}
Accurate prediction of traffic accidents across different times and regions is vital for public safety. However, existing methods face two key challenges: 1) Generalization: Current models rely heavily on manually constructed multi-view structures, like POI distributions and road network densities, which are labor-intensive and difficult to scale across cities. 2) Real-Time Performance: While some methods improve accuracy with complex architectures, they often incur high computational costs, limiting their real-time applicability. To address these challenges, we propose SSL-eKamba, an efficient self-supervised framework for traffic accident prediction. To enhance generalization, we design two self-supervised auxiliary tasks that adaptively improve traffic pattern representation through spatiotemporal discrepancy awareness. For real-time performance, we introduce eKamba, an efficient model that redesigns the Kolmogorov-Arnold Network (KAN) architecture. This involves using learnable univariate functions for input activation and applying a selective mechanism (Selective SSM) to capture multivariate correlations, thereby improving computational efficiency. Extensive experiments on two real-world datasets demonstrate that SSL-eKamba consistently outperforms state-of-the-art baselines. This framework may also offer new insights for other spatiotemporal tasks. Our source code is publicly available at \url{http://github.com/KevinT618/SSL-eKamba}.
\end{abstract}
\begin{IEEEkeywords}
Traffic Accident Forecasting,Spatio-temporal Self-Supervised Learning, Kolmogorov-Arnold, State Space Mode.
\end{IEEEkeywords}

\section{Introduction}
\IEEEPARstart{T}{raffic} accidents represent a significant global public safety issue. With the ongoing urbanization and the increasing number of motor vehicles, both the frequency and severity of traffic accidents continue to rise. According to the World Health Organization (WHO) statistics from 2018, more than 1.3 million people lose their lives in traffic accidents annually, while millions more suffer injuries. This results in substantial losses for society and families, as well as significant economic burdens for governments worldwide. Although traffic accidents exhibit high levels of randomness and complexity, analysis of historical data can reveal underlying patterns and influencing factors. Consequently, traffic accident prediction has become a crucial area of research in traffic safety. Effective prediction can assist governments and traffic management authorities in taking preventive measures in advance, thereby reducing the likelihood and impact of accidents. Additionally, location-based services (such as navigation applications) can leverage this information to provide users with safer routes by understanding dynamic traffic accident risks.

Over the past few decades, statistical methods and machine learning approaches\cite{abdel2004linking,lee2003real,pande2005freeway,hossain2012bayesian,yu2013utilizing}, such as Support Vector Machines (SVMs) \cite{chen2015forecasting} and Random Forests \cite{liu2017prediction}, have been employed to develop traffic accident prediction models for highways and major urban roads. However, these models often fall short in practical applications due to their limited nonlinear representation capabilities. In recent years, with advancements in deep learning technology, both the nonlinear representation capabilities and the ability to handle large-scale data have significantly improved. Researchers have utilized deep learning models such as Graph Convolutional Networks (GCNs)\cite{kipf2016semi}, Long Short-Term Memory networks (LSTMs)\cite{hochreiter1997long}, and Transformers\cite{vaswani2017attention} to more effectively extract complex features from traffic accident data, achieving better performance than traditional methods.

However, these research methods are primarily limited to single road segments and fail to effectively capture the complex spatiotemporal correlations of traffic accidents across different regions within a city \cite{li2020application,theofilatos2017incorporating,yuan2018utilizing,
barba2014smoothing}. Additionally, these models typically involve high computational costs, making them less suitable for traffic tasks that require high real-time performance. Some recent studies\cite{wang2021traffic,trirat2023mg,diao2023dmstg} have attempted to model global spatiotemporal heterogeneity by adopting multi-view graph convolutional networks.

However, the effectiveness of these models largely depends on manually collected spatiotemporal features of regions, such as nearby Points of Interest (POI) and the density of the road network. More importantly, this dependency limits the generalization capability of the models, meaning that they may not effectively capture and predict traffic accident variations when applied to unseen regions or new urban environments. This limitation poses significant challenges for the widespread adoption and application of these models in practical scenarios. Additionally, there is a lack of methods that effectively improve computational efficiency for accident prediction. This paper focuses on how to adaptively capture spatiotemporal heterogeneity across different regions within various cities for traffic accident prediction while maintaining high computational efficiency. Based on this, we also emphasize two main challenges that this paper aims to address.

\begin{figure}[ht!]
   \setlength{\tabcolsep}{6.8mm}
  \centering
  \includegraphics[width=0.5\textwidth]{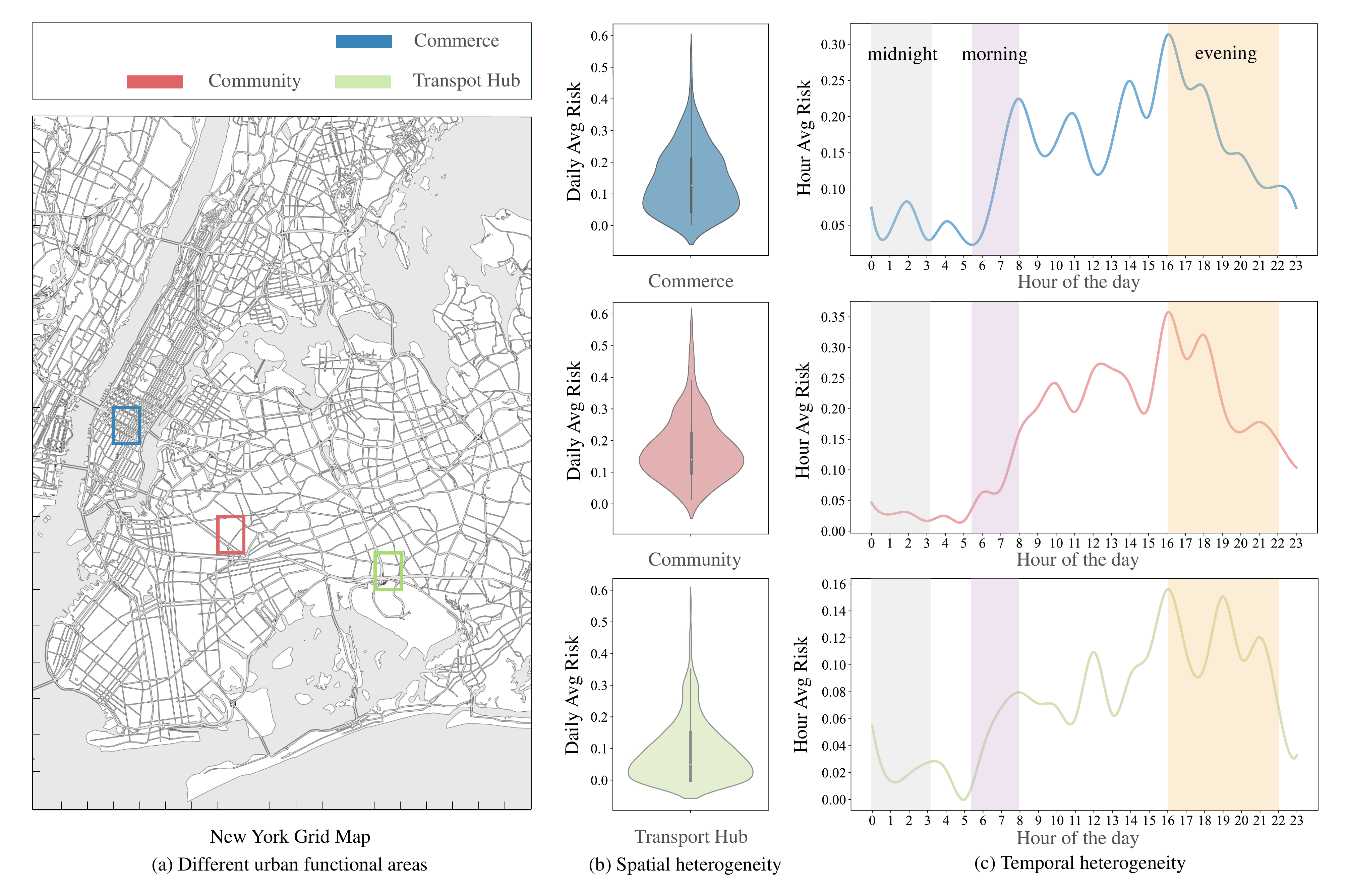} 
  \caption{This underscores the motivation for studying spatio-temporal heterogeneity modeling in traffic accident prediction.
}
   \label{fig1} 
  
\end{figure}

\subsection{Spatio-temporal Variability Modeling}
Fig \ref{fig2}
Urban areas exhibit significant spatio-temporal variability in traffic accident patterns due to differences in Points of Interest (POI) distributions, road characteristics, and functional zoning. For instance, Fig. \ref{fig1} (a) illustrates three distinct areas in New York City, represented by red, blue, and green, corresponding to commercial districts, residential neighborhoods, and transportation hubs, respectively. As depicted in Figure 1b, these areas demonstrate markedly different accident risk distributions. Recent studies \cite{wang2021traffic,khan2019multi,geng2019spatiotemporal,xie2020mgat,sun2020predicting,wang2021gsnet,tamerius2016precipitation,liang2017inferring,chen2016learning} have shown that traffic accidents are influenced by a complex interplay of factors, including both spatially adjacent and contextually related regions. For example, geographically distant residential areas may exhibit similar traffic accident patterns. However, many existing models struggle to effectively capture this spatial variability.Although some studies have developed multi-view Graph Convolutional Networks (GCNs) to model spatial variability, the generalization capability of these models remains limited. Additionally, traditional traffic accident prediction models [31-33] typically treat all time periods uniformly, applying the same set of parameters for predictions regardless of whether it is morning, evening, or nighttime. This uniform approach fails to accurately capture the global temporal semantic variability, as traffic accident patterns vary significantly across different times of the day. For instance, as illustrated in Figure 1c, the distribution of accidents is typically higher during the morning rush hour due to increased traffic density, often concentrated within specific time windows. Conversely, during nighttime, when traffic volume decreases, the accident distribution becomes more balanced and may exhibit a more dispersed and smoother pattern, particularly during holiday periods.

\subsection{Model Complexity and Performance Tradeoffs}

In time series prediction tasks, including traffic accident prediction, model architectures have progressively evolved, with increasing complexity—from Multi-Layer Perceptrons (MLPs) \cite{rumelhart1986learning} and Recurrent Neural Networks (RNNs) \cite{sherstinsky2020fundamentals} to Transformers and various large-scale models. Despite these advancements, the improvements primarily focus on optimizing neural networks built on the foundation of MLPs. However, the inherent architectural limitations of these models pose challenges for deployment in real-world scenarios that require the simultaneous processing of large datasets or adherence to stringent real-time requirements. For instance, MLPs exhibit insufficient nonlinear representation capabilities and necessitate a substantial number of parameters and multiple layers when dealing with high-dimensional features, leading to inefficiencies and potential gradient explosions. The Transformer's self-attention mechanism \cite{vaswani2017attention}, with its quadratic computational complexity, results in a significant increase in computational overhead as the number of input variables or the backtracking length grows. Consequently, in real-time prediction tasks, Transformers may face performance bottlenecks due to high computational and memory demands, particularly in resource-constrained environments.

The recently proposed state space model, Mamba \cite{gu2023mamba}, introduces a selectivity mechanism that allows the model to reason selectively about inputs. Although the model's parameters remain unchanged during inference, this mechanism enables the model to differentiate among various inputs. For instance, the model can concentrate on certain inputs while selectively disregarding others. This approach enhances the model's inference efficiency and flexibility without increasing the parameter count, thereby achieving a balance between performance and computational efficiency to some extent. However, this selective mechanism differs from the Transformer's self-attention mechanism, which calculates the interrelationships among all global variables. To reduce computational overhead, Mamba selectively processes information, allowing it to either focus on or ignore specific inputs (with the focus level defined as context richness). Nevertheless, a lower context richness can impair Mamba's ability to capture complex spatio-temporal dependencies effectively.

To address the aforementioned challenges, we propose an efficient spatio-temporal self-supervised learning framework, SSL-eKamba, for traffic accident risk prediction. The main concepts underlying this framework are as follows:

To tackle the first issue, we introduce a novel self-supervised learning framework designed to adaptively capture spatio-temporal semantic variability. Initially, we perform data and structural augmentation to generate diverse feature representations for the self-supervised task. To effectively capture spatial variability, we utilize a reconstruction learning paradigm that enables the model to learn multiple features from spatial data and form soft clustering structures guided by a K-means objective. This auxiliary self-supervision mechanism enhances the model's ability to sense spatial variability in representations across different regions. For capturing temporal variability, we employ a contrastive learning paradigm. This approach allows the model to identify and characterize temporal dynamic features by comparatively learning the distribution of traffic accidents across various time periods and at specific temporal nodes.

To address the trade-off between model complexity and performance, we explore whether it is possible to improve prediction accuracy in Mamba, even with reduced contextual richness, while maintaining manageable computational overhead. Our approach involves enhancing the nonlinear representations within the encoding layer. Unlike traditional methods, which typically rely on linear weight parameters for backpropagation, we propose a direct nonlinear transformation of the input using a B-spline function. This method eliminates the need for linear weight parameters and instead leverages a nonlinear activation function for direct backpropagation. This strategy is inspired by the recently proposed Kolmogorov-Arnold Network (KAN) architecture, which has demonstrated the effectiveness of nonlinear transformations in complex model architectures.

In summary, the main contributions of this work are as follows:
\begin{itemize}
    \item 
Cross-Domain Generalization: We introduce a novel self-supervised learning framework to capture spatio-temporal variability across regions, using contrastive learning for traffic time dynamics and reconstruction learning for spatial variability.
\item 
Addressing Data Sparsity: We mitigate the effects of data sparsity by incorporating additional self-supervised signals, improving model performance in sparse data scenarios.
\item 
Efficient State Space Model: We propose SSL-eKamba, a state-space model that balances efficiency and effectiveness by using Kolmogorov-Arnold theory and spatio-temporal self-supervised training, demonstrating robust traffic accident prediction across different cities.
\item 
Extensive Validation: Our extensive evaluations on two real-world datasets show that SSL-eKamba outperforms existing models, with potential applications in other domains like power demand forecasting and logistics computing.
\end{itemize}

The remainder of this paper is organized as follows: Sec.\ref{sec2} reviews related work. Sec.\ref{sec3} introduces the preliminary concepts and provides a detailed explanation of the proposed model’s structure. Sec.\ref{sec4} presents the experimental results. Finally, Sec.\ref{sec5} concludes the paper and discusses potential future research directions.

\section{RELATED WORK}
\label{sec2}
We provide an overview of related work in the field of traffic accident prediction, with a focus on discussing the efficiency challenges in real-time traffic accident forecasting. Finally, we summarize the relevant work on self-supervised learning 

\subsection{A.Traffic Accident Prediction}
Over the past few decades, significant advancements have been made in predictive modeling of traffic accident risks and in improving traffic safety \cite{abdel2004linking,lee2003real,pande2005freeway,hossain2012bayesian,yu2013utilizing,chen2015forecasting,liu2017prediction}. Early pioneering research primarily utilized statistical methods and machine learning-based approaches to develop traffic accident prediction models for highways and urban arterial road segments \cite{kipf2016semi,hochreiter1997long,vaswani2017attention,li2020application,theofilatos2017incorporating}. Some of these early approaches employed linear models to analyze traffic accidents across multiple road sections. More recently, the scope of traffic accident risk assessment has been expanded from individual urban arterial roads to entire urban areas by dividing these areas into regular grids [citation needed]. However, these studies often rely on traditional methods applied to small-scale traffic accident datasets with limited features, which are inadequate for effectively capturing the complex spatio-temporal dependencies inherent in the data.

In recent years, deep learning-based methods have demonstrated superior capabilities in representing nonlinear relationships and learning spatio-temporal patterns from contextual data. Recent studies \cite{10023949} have started to jointly address temporal and spatial dependencies in traffic-related data using deep learning models. For instance, Ren et al. extended Recurrent Neural Network (RNN) architectures and fully connected (FC) networks to capture temporal dynamics for predicting future traffic accident risks. In a more advanced approach, Yuan et al. proposed a heteroconvolutional long- and short-term memory (HeteroConvLSTM) model to capture both spatial heterogeneity and temporal autocorrelation of traffic accidents. Similarly, Bao et al. applied a spatio-temporal convolutional LSTM network to predict short-term crash risks in New York City at an hourly level \cite{wang2021gsnet}. However, these approaches often overlook the differential impact of various external semantic features.

Graph-based Convolutional Networks (GCNs) have proven to be effective in the field of traffic prediction, with several researchers \cite{khan2019multi,geng2019spatiotemporal,xie2020mgat,sun2020predicting,wang2021gsnet}[riskseq, vit, gumt] employing GCNs for multi-scale prediction tasks. Emerging research has designed methods to capture external semantic correlations by manually constructing semantic graphs of weather, Points of Interest (POIs), mobility, and traffic accidents. These methods aim to learn semantic and geographic features, respectively, and utilize an attention module to measure the influence weights of different view features [gsnet, mgtar, mvmt-stn].

In contrast to these approaches, our work introduces a spatio-temporal self-supervised learning framework that optimizes the process by adaptively capturing the variability of various external semantic features. This approach significantly enhances the model's generalization capability, providing a more robust solution to the complexities of traffic prediction.

\subsection{Real-time Performance Issues}
Given the large volumes of complex spatio-temporal data in real-world traffic scenarios, models inevitably face significant computational overhead. Most previous approaches have prioritized prediction performance while overlooking the computational challenges that arise during deployment. This issue can be mitigated through two primary strategies: reducing the number of model parameters or optimizing model efficiency through various techniques.

One approach involves reducing model parameters using strategies such as pruning techniques and deep compression. \cite{alqahtani2021pruning} explored filter pruning in convolutional layers to decrease the number of parameters and reduce the model's computational cost by eliminating less important filters. \cite{liu2017learning} proposed network thinning, where channels in convolutional networks are pruned by introducing sparsity metrics during training, effectively lowering the model's complexity and memory requirements. Additionally, Song Han et al. introduced a deep compression technique that integrates pruning, quantization during training, and Huffman coding. This method significantly reduces the storage requirements of neural networks and enhances computational efficiency. It has been shown to substantially decrease model storage space and accelerate the inference process, making it a crucial tool for facilitating model deployment.

Another approach to improving computational efficiency is the use of parallel training. \cite{seide20141} explored the impact of synchronous and asynchronous updating strategies on training effectiveness in data-parallel environments. They proposed techniques such as delayed updating and reducing the communication overhead, which significantly enhanced the efficiency of data parallelism in multi-GPU settings. These methods contribute to reducing synchronization overhead and optimizing parallel processing, making data parallelism more effective.

In addition to parallel training strategies, innovative network architectures \cite{wen2020fast} can also play a crucial role in improving efficiency. A notable example is the recently proposed Mamba state-space model \cite{gu2023mamba}, which introduces a selectivity mechanism through parameterized learning during the training phase. This mechanism enhances the model's inference efficiency without increasing the number of parameters, providing a streamlined solution to the computational challenges posed by complex models.

\subsection{Self-Supervised Learning}
The aim of self-supervised learning is to obtain labeled signals from unlabeled data through proxy tasks \cite{jing2020self,gao2021simcse,wen2022robust}[Ren et al., 2021; Ren, Wang, Zhao, 2022; Wu et al., 2020]. Many studies have utilized self-supervised learning architectures to enhance the generalization ability of models. Fine-tuning these general representations in downstream subtasks has been shown to significantly improve performance.

\cite{kiros2015skip} employed self-supervised learning to learn representations by predicting contextual sentences based on a given sentence. \cite{gan2016unsupervised,ma2019end} proposed a temporal representation learning model that leverages the reconstruction learning paradigm for time series clustering.

Similarly, in vision tasks\cite{caron2020unsupervised} utilized current clustering results to iteratively update the network, thereby enhancing feature representations, which were subsequently used to optimize clustering assignments. A more prevalent approach \cite{chen2020simple,gao2021simcse,hassani2020contrastive,oord2018representation,wang2020understanding,you2020graph} involves learning by contrasting positive samples with similar features against negative samples with different features. 

Time-level models \cite{wen2020fast,zhou2022film,sagheer2019unsupervised,audibert2020usad,franceschi2019unsupervised} address time series-specific challenges by focusing on scale-invariant representations of individual timestamps. Drawing inspiration from these approaches, our study introduces two novel spatio-temporal self-supervised assisted training schemes, which have not been thoroughly explored in existing traffic accident prediction methods.

\begin{figure}[ht!]
  \centering
  \includegraphics[width=0.5\textwidth]{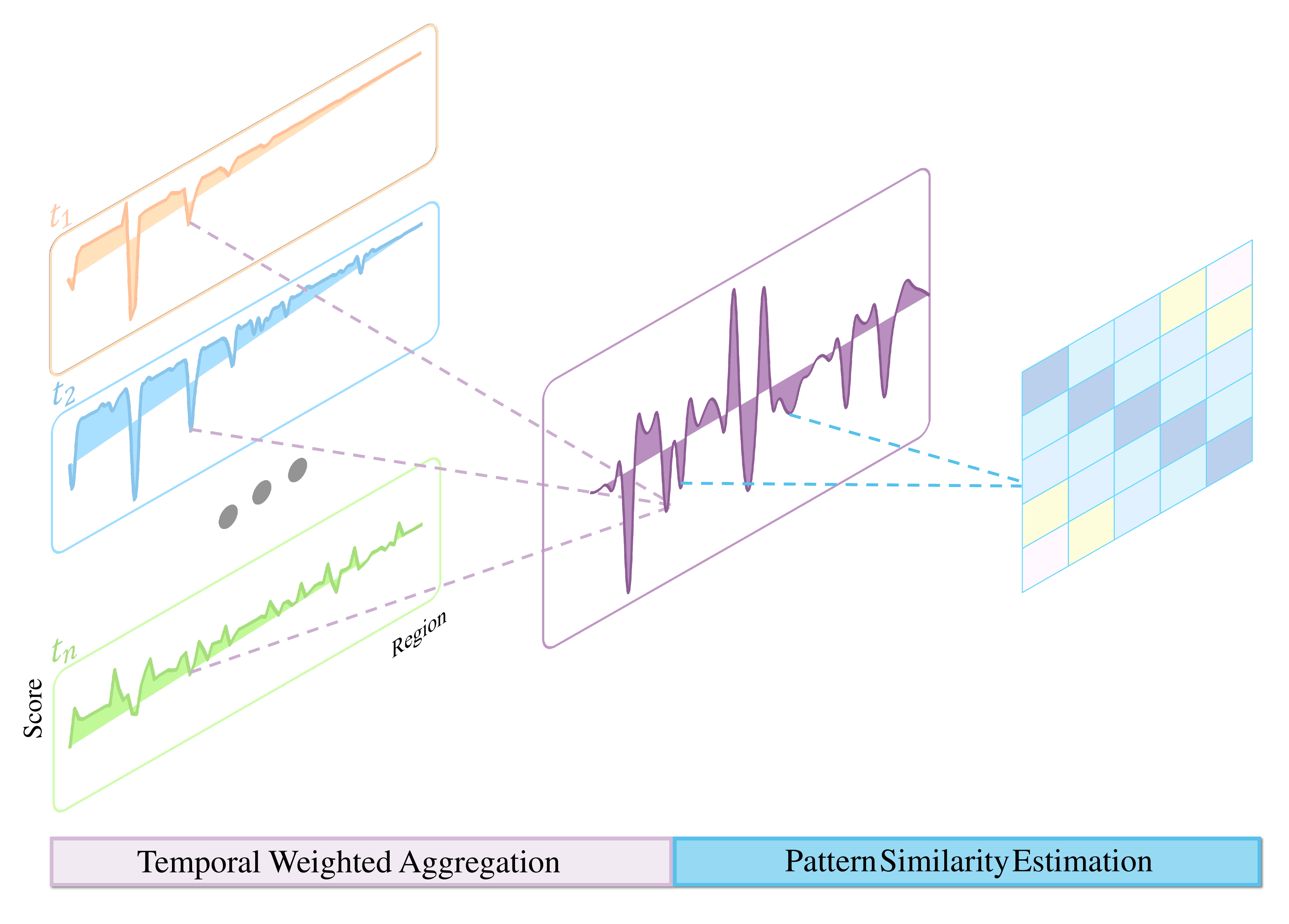} 
  \caption{This underscores the motivation for studying spatiotemporal heterogeneity modeling in traffic accident prediction.
}
  \label{fig3}
\end{figure}
\section{METHOD}
\label{sec3}
In this section, we begin by introducing the problem definition and providing a formal definition of the research problem (Sec. \ref{secA}). Next, we offer a detailed description of the proposed SSL-eKamba framework (Sec. \ref{secB}). Finally, we discuss the training strategy employed in this study (Sec. \ref{secC}).

\subsection{Task Definition}
\label{secA}
\textit{Definition 1 (Spatial Region):} 
A city is divided into $N= I\times J$ disjoint geographic grids based on longitude and latitude, where each cell or grid region is considered a spatial region $r_{n}(1\le n\le N) $.We denote the set of spatial regions in the city as $\mathcal{V}=\{r_{1},\ldots,r_{N}\}$.

\textit{Definition 2 (City Graph):} 
A city graph $\mathcal{G}=\{\mathcal{V},\mathcal{E},\mathcal{A}\}$ is defined as an undirected graph comprising a collection of nodes (city regions). For any two regions $r_{i},r_{j} \in \mathcal{V}$, the edge $e_{i,j}\in \mathcal{E}$,represents the connection between these two neighboring regions. The adjacency matrix corresponding to this graph can be defined as follows:$$\mathcal{A}_{i,j}=\begin{cases}1,&v_i\text{ to } v_j\text{ are connected}\\0,&v_i \text { to }  v_j \text{ are not connected}\end{cases}$$

\textit{Definition 3 (Human Activity Tensor):} 
Human activity data, such as taxi trip data, for all cells in each time slot ${t}$ is represented as a tensor $S_{t}^{} \in \mathbb{R} _{}^{I\times J\times 2} $.

\textit{Definition 4 (Traffic Accident Risk):} 
Traffic accidents are classified into three types based on the number of casualties: minor accidents, injury accidents, and fatal accidents, which are assigned risk values of 1, 2 and 3 respectively. We denote $Risk_{t}^{} \in \mathbb{R} _{}^{I\times J} $as the risk map at time slot ${t}$, where each entry $Risk_{t,i}^{}  $ represents the sum of risk values for the region $r_{i} \in \mathcal{V}$ at time ${t}$. This term serves as a proxy for the risk level of traffic accidents.

\textit{Problem Statement:} 
Let $X_{t}^{} \in \mathbb{R} _{}^{I\times J\times d} $ represent the grid features, including traffic accident risk, traffic flow, and other relevant information for all regions at the current time step ${t}$,as well as the spatial region $\mathcal{V}$ and city graph $\mathcal{G}$ of a given city. The objective is to learn a prediction function that can efficiently predict the traffic accident risk map $Risk_{t+1}^{}\in \mathbb{R} _{}^{I\times J}$ at the next time interval.

\subsection{Model Architecture}
\label{secB}
As illustrated in Fig \ref{fig2}, the SSL-eKamba model follows the paradigm of Data Enhancement → Spatio-Temporal Encoding → Self-Supervised heterogeneous Learning → Prediction. The detailed components of this architecture are described below.

\subsubsection{\textbf{Adaptive Data Augmentation}}
In this subsection, we focus on adaptive data augmentation applied to the output $(C_{t-T }^{},\ldots,C_{t}^{})$ from the first time encoder. Previous work \cite{yuan2019arterial,bao2019spatiotemporal} has shown that regions with similar semantic information often exhibit homogeneous data distributions, such as similar traffic patterns. Inspired by this observation, we design two data augmentation schemes to learn the spatio-temporal variability between regions. These schemes are based on a heterogeneity metric and include accident-level data augmentation and graph-structure-level augmentation. The proposed augmentation strategies adaptively identify potential accident risk data and mitigate bias from low-correlation traffic patterns in connected regions, while also enhancing the model’s ability to capture dependencies between distant regions.

\textbf{Local-to-Global Discrepancy}. As illustrated in Figure 3a, for a region $r_{n}$ at a specific time step ${t}$, the embedding sequence $(c_{t-T,n }^{},\ldots,c_{t,n }^{})$ represents the sequence of embeddings within a ${T}$-step window $(C_{t-T }^{},\ldots,C_{t}^{})$ in a single row. This sequence is used to generate the variability score between the regional traffic pattern and the overall traffic pattern at time step ${t}$ follows:
\begin{equation}
q_{\varphi  ,n }^{}= c_{\varphi  ,n }^{\top }\cdot w_{0}^{}
\end{equation}
where $\varphi$  is an index for the time step range ${(t-T,t)}$, $ c_{\varphi  ,n }$ is the embedding of region $r_{n}$ at time step ${t}$, and $w_{0}^{}\in \mathbb{R} _{}^{d}$ is a learnable parameter vector for the transformation.As shown in Figure 3b, we can obtain a time-based representation by performing temporal weight aggregation as follows:
\begin{equation}
p_{n}^{}= \sum_{\varphi =t-T}^{t} q_{\varphi ,n}^{} \cdot c_{\varphi,n }^{}
\end{equation}
where $p_{n}^{}$ is the aggregated representation based on the derived weights $q_{\varphi ,n}^{}$ applied to the sequence of embeddings for region $r_{n}$ within the input time steps.

\textbf{Local-to-Local Discrepancy}. To assess the degree of region-to-region heterogeneity and the differences in traffic patterns between regions over time, we employ Pearson's correlation coefficient\cite{cohen2009pearson}. This metric evaluates the correlation between the traffic patterns of different regions, allowing us to quantify their temporal discrepancies as follows:
\begin{equation}
o_{m,n}=\frac{\sum_{\varphi }^t(p_{\varphi,m}-\overline{p_m})(p_{\varphi,n}-\overline{p_n})}{\sqrt{\sum_{\varphi }^t(p_{\varphi,m}-\overline{p_m})^2\sum_{\varphi }^t(p_{\varphi,n}-\overline{p_n})^2}}
\end{equation}
When the value of $o_{m,n}$ is close to 1, it indicates a strong positive correlation in traffic patterns between regions $r_{m}$ and $r_{n}$, suggesting a low degree of heterogeneity between these regions. Conversely, a smaller $o_{m,n}$ means that regions show negative correlation with each other.

\textbf{Incident-Level Data Augmentation}. This approach aims to enhance the data by injecting simulated contingencies into specific regions, inspired by recent advancements in data enhancement strategies [citation needed]. We design an augmentation operator for the traffic tensor $\mathcal{X}_{t-T:t }^{}  $ that adaptively learns the dependencies between regional traffic patterns and the overall city. Specifically, using the local-global similarity score $q_{\varphi  ,n }^{}$, computed earlier, we apply an exponential transformation \textit{i.e,} $\alpha _{\varphi ,n}\sim \frac{\exp(-p_{\varphi ,n} )}{\sum_{i=1}^N\exp(-p_{\varphi ,i})}$ to adjust the probability of injection for each region. Beacuse a region exhibits a lower similarity score, indicating that its traffic pattern is more divergent from the city's overall pattern at time ${t}$, it may suggest a higher hidden risk of accidents. The augmented data with the incident-
level augmentation is denoted as $\tilde{\mathcal{X}}_{t-T:t}$.

\textbf{Graph Structure-Level Augmentation}. We perform graph topology enhancement on $\mathcal{G}$, guided by the heterogeneity metric $o_{m,n}$. Specifically, for adjacent regions $r_{m}$ and $r_{n}$, if their heterogeneity is high, we apply a masking operation to the edge $e_{i,j}\in \mathcal{E}$ based on the mask probability derived from the exponential transformation \textit{i.e,} $\beta_{m,n}\sim\frac{\exp(-o_{m,n})}{\sum_{i=1}^{I}\sum_{j=1}^{J}\exp(-o_{m,(i,j)})}$. Conversely, for two non-adjacent regions with a low discrepancy score, we add a new edge between them.The augmented data with the graph-
level augmentation is denoted as $\tilde{\mathcal{G}}$.

\subsubsection{\textbf{Temporal Embedding Layer}}
According to previous research \cite{xu2016big}, traffic conditions exhibit significant temporal periodicity, with distinct patterns emerging across various time periods. Traffic accidents are often influenced by the recent traffic conditions in a given region. Therefore, capturing temporal dependencies is crucial for accurate prediction. As discussed in the introduction, the Selective State Space Model (SSM) \cite{gu2023mamba} has lower contextual richness compared to attention mechanisms, making it more challenging to extract complex dependencies in traffic data. To address this limitation, we introduce a new model, eKamba, specifically designed to learn the temporal dependencies of historical traffic accident risks.

We employ the Efficient Kolmogorov-Arnold Network (eKAN) for nonlinear projection and residual connectivity. Unlike traditional neural networks that utilize fixed activation functions (such as those in Multi-Layer Perceptrons, or MLPs), eKAN introduces learnable activation functions positioned on the edges of the network. This innovative design allows each weight parameter in the KAN \cite{liu2024kan} to be substituted with a univariate function, often parameterized as a spline function. This approach enables the model to process higher-dimensional feature spaces, thereby capturing more complex and nuanced features within the data.

It is important to highlight that in KAN, the activation function for each input is formed by a linear combination of a set of basis functions, which necessitates extending the dimensionality of the input. This extension significantly increases computational complexity.

To address this efficiency issue, we have reformulated the computational architecture of KAN by utilizing different basis functions to activate the inputs, which are then linearly combined. In practice, eKAN first computes the sequence data $X_{t}^{} $, derived from the most recent $\rho $ time intervals and the corresponding time intervals from the previous $\kappa $ weeks. This process generates the initial feature mapping, which serves as the foundation for further processing in the network.
The output of eKAN can be derived as:
\begin{equation}
e(x)=\sum_{k=1}^Kw_k\cdot B_k(x)
\end{equation}
\begin{equation}
\Phi(x)=W_\text{base}\cdot\sigma(x)+W_\text{spline}\cdot e(x)+b
\end{equation}
\begin{equation}
H_{}^{\prime }=eKAN(x)=\Phi_L\circ\Phi_{L-1}\circ\cdots\circ\Phi_0(x)
\end{equation}

where $W_\text{base}$ and $W_\text{spline}$  the weight matrices associated with the spline and basis functions, respectively. ${b}$ represents the bias vector of the linear layer. $\alpha(x)$ is the residual activation function, specifically the SiLU (Sigmoid Linear Unit) function, which acts as a jump connection similar to those in a residual network (ResNet). $B_k(x)$ denotes the basis functions, specifically B-spline functions, that are used to activate the input features $X_{t}^{}$. $w_k$ is the weight associated with each basis function $B_k(x)$.
Based on this, the Temporal Embedding Layer can be expressed as:
\begin{equation}
H=SiLU(SSSM(Conv1D(H_{}^{\prime } )))+SiLU(H_{}^{\prime } )
\end{equation}
\begin{equation}
H=eKAN(H)
\end{equation}

\begin{figure*}[ht!]\label{fig2}
  \centering
  \includegraphics[width=1\textwidth]{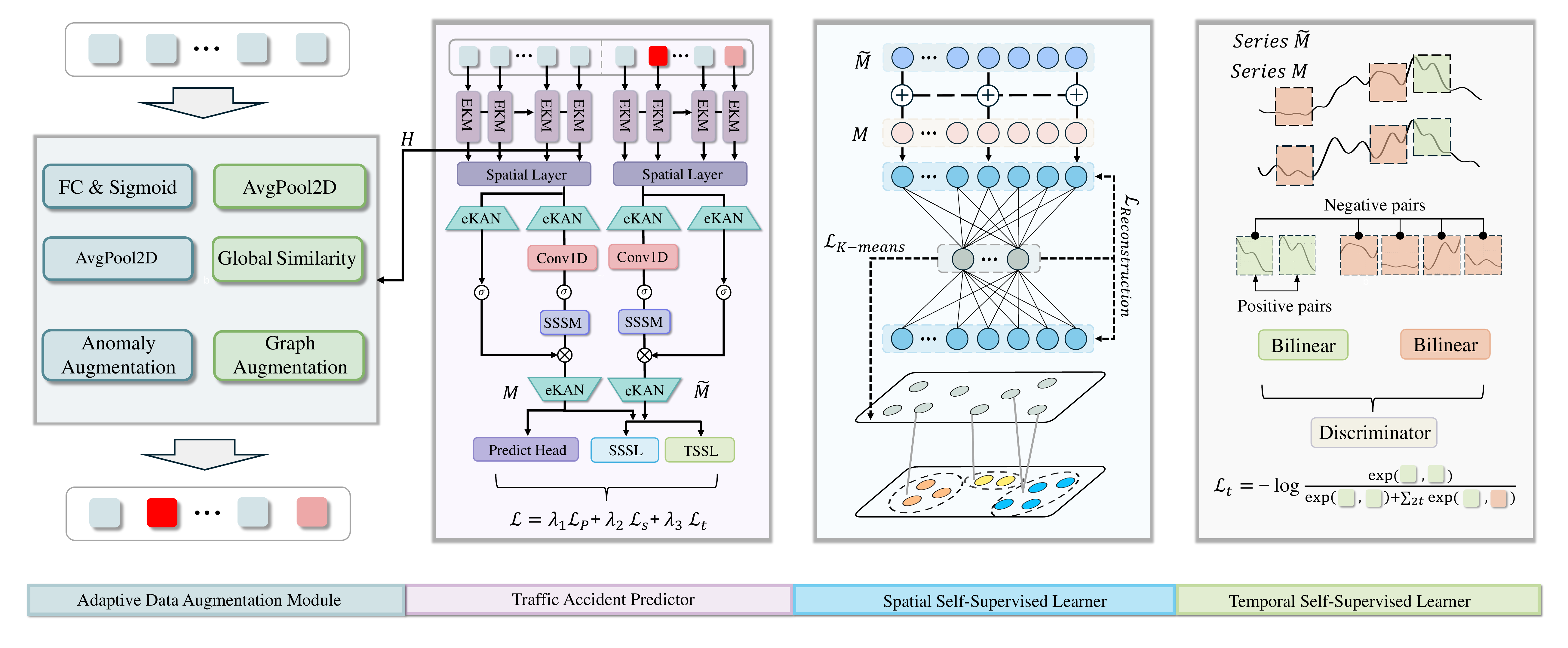} 
  \caption{This underscores the motivation for studying spatiotemporal heterogeneity modeling in traffic accident prediction.
}
  \label{fig2}
\end{figure*}
\subsubsection{\textbf{Spatial Embedding Layer}}
The entire urban traffic network can be represented as a graph structure, where regions connected by roads define the elements of the adjacency matrix ${A}$ as 1; otherwise, the elements are set to 0. To capture the spatial dependencies between these regions, we employ a Graph Convolutional Network (GCN) \cite{kipf2016semi,zhang2019graph}, a model that has been validated in numerous previous studies [citations needed]. For example, for the first layer, the Graph Convolutional Layer Network takes as input the city graph $\mathcal{G}$, and the time-aware embedding $H\in\mathbb{R}^{N\times D}$. Following the standard form of GCN processing, we reorganize the matrix as follows:
\begin{equation}
\hat{A}=\tilde{D}^{-\frac12}\tilde{A}\tilde{D}^{-\frac12}
\end{equation}

where $\tilde{A}=A+I_N$ is the adjacency matrix with self-connections, and $\tilde{D}_{ii}=\sum_j\tilde{A}_{ij}$ is the degree matrix corresponding to $\tilde{A}$.Given this information and a set of trainable weights $W$, at time interval $t$, a layer of the Graph Convolutional Network (GCN) can be modeled as: 
\begin{equation}
H_{t}^{(l+1)}=\mathrm{ReLU}({\hat A}^{(l)}H_t^{(l)}W_t^{(l)})
\end{equation}

where $H_t^{(l)}$ indicates the hidden representation of the graph convolution at the $l$-th layer, $W_t^{(l)}$ denotes the weights of the corresponding convolution kernels in layer $l$.

Our spatio-temporal encoder is designed using a layered structure, where each layer is composed of a combination of eKamba, GCN, and eKamba layers, as illustrated in Figure 3b. This layered structure is stacked $L$ times to construct the complete STencoder model. Through embedding propagation and feature aggregation, the model produces an embedding matrix $M\in\mathbb{R}^{N\times D}$. Each row vector $m_n\in\mathbb{R}^{D}$ of the matrix $M$ corresponds to the region $r_n$ and represents the final embedding of that region. 

\subsubsection{\textbf{Self-Supervised Spatial Disparity Learning}}                             

We adopt the concept of deep time series representation clustering [citation needed] and extend it to a multi-region time series clustering model. This approach maps regions into multiple potential representation spaces that correspond to different functional urban areas (e.g., residential areas, shopping centers, transportation hubs). The objective is to ensure that the region embeddings effectively preserve spatial heterogeneity by leveraging an auxiliary self-supervised signal.Specifically, spatial dissimilarity learning is divided into two key components: \textbf{reconstruction loss learning} \cite{baldi2012autoencoders,abid2018autowarp}and \textbf{K-means loss-guided representation learning} \cite{petitjean2011global,aghabozorgi2015time} . In the first part, the input data is compressed into a low-dimensional latent representation. This compression aims to capture the essential features and underlying structure of the data by attempting to reconstruct the original input from this latent space. The goal is to learn a latent representation that reveals the intrinsic characteristics and structure of the data. In the second part, k-means clustering is integrated into the model to guide the representation learning process. This integration helps to align the learned representations with the underlying data distribution, enhancing the model's ability to capture meaningful patterns and clusters within the data. The following sections will provide a detailed explanation of these processes.

We firstly fuse the original data with the augmented data to obtain :
\begin{equation}
v_{n}=\boldsymbol{w}_{1}\odot\boldsymbol{m}_{n}+\boldsymbol{w}_{2}\odot\tilde{\boldsymbol{m }}_{n}
\end{equation}
where $v_{n}$ represents the embeddings of region $r_n$, $\odot$ is the element-wise product. $\boldsymbol{w}_{1}$,$\boldsymbol{w}_{2}$ are learnable parameters. After that,We define a linear mapping $f_{enc}:v\to D_{n}^{}$ and $f_{dnc}:D_{n,k}^{}\to v_{}^{\prime} $, $f_{enc}$ and $f_{dnc}$ represent the encoding and decoding processes, respectively. Here, $D_n\in\mathbb{R}^{d\times N}$ denotes the d-dimensional latent representation the embedding of $v_{n}$, defined as follows:
\begin{equation}
\begin{cases}
D=f_{enc}^{} (v)
 \\
v_{}^{\prime } =f_{dnc}^{} (D)
\end{cases}
\end{equation}
Our objective is to train a robust function $f$ such that the learned representation is conducive to heterogeneity-aware region clustering. 
To achieve this, we employ Mean Squared Error (MSE) as the reconstruction loss, defined as follows:
\begin{equation}
\mathcal{L}_{\text{reconstruction}}=\frac{1}{n}\sum_{i=1}^{n}\|v_{i}-\hat{v}_{i}\|_{2}^{2}
\label{equ13}
\end{equation}
It is important to note that while the representations learnt from the reconstruction loss capture the underlying structure and features in the data, they are not necessarily suitable for heterogeneity-aware region clustering. To address this, we introduce a regularization objective based on K-means clustering during optimization. \cite{zha2001spectral} demonstrated that the K-means objective can be reformulated as a trace maximization problem associated with the Gram matrix $D^\top D$. Spectral relaxation converts the
K-means objective into the following problem:
\begin{equation}
\mathcal{L}_{\text{K-means}}=\mathrm{Tr}(D^\top D)-\mathrm{Tr}(F^\top D^\top DF)
\label{equ14}
\end{equation}

where $\mathrm{Tr}$ denotes the matrix trace. $F\in\mathbb{R}^{N\times k}$ is the cluster indicator matrix.

\subsubsection{\textbf{Self-Supervised Temporal Disparity Learning}}    

In this component, we design a self-supervised task to introduce temporal heterogeneity into time-aware region embeddings by learning the disparity between traffic pattern representations across different time steps.

Specifically, we generate positive-negative sample pairs from the region embedding $m_n$ of the region $r_n$obtained by encoding the original data, and the embedding $\tilde{\boldsymbol{m }}_{n}$ of the region $r_n$ obtained by encoding the augmented data. To enhance the distinction between representations at different time steps, we consider the same region at the same time step as positive sample pairs $({m}_{t,n},\tilde{m}_{t,n})$, while embeddings from the same region at different time steps are treated as negative sample pairs $({m}_{t,n},\tilde{m}_{t_{}^{\prime } ,n})$.The similarity of positive and negative sample pairs is calculated as follows:
\begin{equation}
z_{t,n}^{\prime } ={Bilinear}({m}_{t,n},\tilde{m}_{t,n})
\end{equation}
\begin{equation}
z_{t,n}^{\prime \prime } ={Bilinear}({m}_{t,n},\tilde{m}_{t_{}^{\prime } ,n})
\end{equation}
where $t$ and $t_{}^{\prime }$ denote two different time steps.Formally, the temporal heterogeneity self-supervised task is optimized through the following loss function:
\begin{equation}
\ell  =-\log\frac{\exp(\mathbf{z}_{t,n}^{\prime}/\tau)}{\exp(\mathbf{z}_{t,n}^{\prime}/\tau)+\sum_{j}^{}\exp(\mathbf{z}_{j,n}^{\prime\prime}/\tau)}
\end{equation}
where $\tau$ is the temperature parameter to control the smooth-
ing degree of softmax output. The overall self-supervised
objective over all reigons is defined as follows:
\begin{equation}
\mathcal{L}_{t}=\frac{1}{N}\sum_{n=1}^{N}\ell (z_{t,n}^{\prime },z_{t,n}^{\prime \prime })
\label{equ18}
\end{equation}
This design allows positive sample pairs to reinforce consistency in traffic patterns at a given time (e.g., during rush hour or under specific weather conditions), while negative sample pairs help capture temporal variability across different time steps, enriching the model's ability to learn dynamic traffic trends.
\subsubsection{\textbf{PredictHead}}
To map the embedded representation $m_n$ of each region $r_n$ learned by SSL-eKamba into a more separable feature space, we employ an MLP to estimate the accident risk score for each region $r_n$.
The traffic accident prediction for the future time step $t+1$ is given by:
\begin{equation}
    \hat{\boldsymbol{x}}_{t+1,n}=\mathrm{MLP}(m_n)
\end{equation}
where $\hat{\boldsymbol{x}}_{t+1,n}$ represents the predicted accident risk for region $r_n$ at time $t+1$.

\subsection{{Training Strategy}}    
Due to the rarity of traffic accidents, the model's prediction loss function must be carefully designed. For instance, using absolute error loss for training tends to bias the model's predictions towards 0, thereby minimizing the loss but resulting in predictions of almost no accidents. On the other hand, using squared error loss makes the model highly sensitive to outliers. To address these issues, we employ a weighted loss function \cite{wang2021gsnet}, where samples with higher traffic accident risk are assigned greater weight during training, thereby avoiding a scenario where the final prediction is zero.Specifically, we categorize all samples into four risk levels based on accident risk, denoted as $i= \left \{ 0,1,2\ge 3  \right \} $. The final prediction loss can be expressed as:
\begin{equation}
    \mathcal{L}_p({X},\hat{{X}})=\frac12\sum_{i\in{I}}\lambda_i({X}(i)-\hat{{X}}(i))^2
\end{equation}
where $X$ represents the sample with a traffic accident risk level of $i$, ${X},\hat{{X}}$ denote the predicted and actual values, respectively, and $\lambda_i$ is the risk weight hyperparameter.

Finally, by incorporating the self-supervised spatio-temporal heterogeneity learning losses from Equations (\ref{equ13}), (\ref{equ14}) and (\ref{equ18}) into the joint learning objective, the overall training loss of the model is defined as:
\begin{equation}
    \mathcal{L}= \lambda _1\mathcal{L}_{p}+\lambda _2\mathcal{L}_{s}+\lambda _3\mathcal{L}_{t}
\end{equation}
where $\mathcal{L}_{s}=\mathcal{L}_{reconstruction}+\lambda _4\mathcal{L}_{K-means}$, $\lambda _{1-4}$ are trade-off hyperparameters.

Our model is trained using the backpropagation algorithm through four stages: First, the spatial and temporal encoder generates a region embedding matrix $M$ from the ($\mathcal{G}$, $\mathcal{X}$). Meanwhile,adaptive augmentation refines ($\mathcal{G}$, $\mathcal{X}$) into ($\tilde{\mathcal{G}}$, $\tilde{\mathcal{X}}$), which are re-encoded by the shared ST encoder to produce $\tilde{M}$. We then compute $\mathcal{L}_s$, $\mathcal{L}_t$ and $\mathcal{L}_p$ losses using ($M$,$\tilde{M}$), and combine them into the joint loss  $\mathcal{L}$. Finally,the model is trained until  $\mathcal{L}$ converges.

\label{secC}
\section{EXPERIMENTS}
\label{sec4}
In this section, we first describe the dataset and evaluation protocol used in the experiments (Sec. \ref{sec_e_A}). We then provide details of our implementation and report the performance of SSL-eKamba across various metrics (Sec. \ref{sec_e_B}). Following this, we present ablation studies and conduct in-depth analyses to assess the contributions of different components (Sec. \ref{sec_e_C}). Finally, we offer qualitative results to further illustrate the model's effectiveness (Sec. \ref{sec_e_D}).
\subsection{{Experimental Setup}}
\label{sec_e_A}
\subsubsection{\textbf{Dataset}}
Table \ref{table1} provides the dataset statistics, where we utilize two publicly available large-scale traffic accident datasets: one from New York (January 2013 to December 2013)\footnote{\href{https://opendata.cityofnewyork.us}{https://opendata.cityofnewyork.us}} and another from Chicago (February 2016 to September 2016)\footnote{\href{https://data.cityofchicago.org}{https://data.cityofchicago.org}}. The road condition information includes average road speed, road volume, crash rate, road width, road direction, number of lanes, snow removal priority, road type, and traffic direction. The traffic accident information encompasses location, time, and the number of casualties.

Notably, our model does not explicitly incorporate graph structures such as Points of Interest (POIs), road features, or demographic data. Instead, it leverages two self-supervised auxiliary tasks to adaptively learn these feature differences.

We divide each city into small areas, with each area approximately 2 km × 2 km. For both datasets, the model uses data from the previous 3 hours near the prediction time, along with data from the previous 4 weeks, to predict accidents at the next time step. A sliding window strategy is employed to generate samples, and the datasets are split into training, testing, and validation sets in a 6:2:2 ratio.

\begin{table}[]
\setlength{\tabcolsep}{5.9mm}
\caption{STATISTICS AND DESCRIPTION OF THE TWO DATASETS}
    \label{table1}
\begin{tabular}{@{}ccc@{}}
\toprule
Dataset           & New York City       & Chicago            \\ \midrule
Time Span         & 1/1/2013-12/31/2013 & 2/1/2016-9/30/2016 \\
Traffic accidents & 147k                & 44k                \\
Taxi Trips        & 173179k             & 1744k              \\
POIs              & 15625               & None               \\
Weathers          & 8760                & 5832               \\
Road Network      & 103k                & 56k                \\ \bottomrule
\end{tabular}
\end{table}
\subsubsection{\textbf{Evaluation Protocol}}
We evaluate the SSL-eKamba model from three perspectives, similar to those commonly used in recent studies [citation needed]:

Regression Perspective: We use Root Mean Square Error (RMSE) to assess the error in predicting risk across all regions. This metric quantifies the model's accuracy in estimating the overall risk levels.

Classification Perspective: We employ Recall@k and Mean Average Precision at k (MAP@k) to evaluate the model's performance in predicting high-risk regions. Recall@k measures the overlap between the predicted high-risk regions and the regions where high-risk events actually occur. MAP@k denotes the average prediction accuracy within the k highest-risk regions, providing insight into how well the model identifies the most critical areas.

Efficiency Perspective: While metrics like RMSE and MAP@k are important, they do not account for the real-time requirements crucial in traffic accident scenarios. Therefore, we also evaluate the model's efficiency in terms of memory usage and computation time, determining whether it meets the operational demands of real-world applications.
\begin{equation}
    \mathrm{RMSE}=\sqrt{\frac{1}{T}\sum_{t=1}^{T}\left(X_{t}-\hat{X}_{t}\right)^{2}}
\end{equation}
\begin{equation}
    \mathrm{Recall}=\frac{1}{T}\sum_{t=1}^{T}\frac{R_{t}\cap R_{t}^{\prime } }{|R_{t}^{\prime}|}
\end{equation}
\begin{equation}
    \mathrm{MAP}=\frac{1}{T}\sum_{t=1}^{T}\frac{\sum_{j=1}^{|R_{t}^{\prime}|}pre(j)\times rel(j)}{|R_{t}^{\prime}|}
\end{equation}
where $X_{t}$ is the ground truth, $\hat{X}_{t}$ is the prediction in time slot $t$, $R_t$ and $R_{t}^{\prime }$ are the sets of actual and predicted top $K=|R_{t}^{\prime}|$ highest risk districts, respectively. $pre(j)$ from rank 1 to $j$ in the list, $rel(j) = 1$ indicates that a traffic accident actually occurred in region $j$ at that time; otherwise $rel(j) = 0$.

\subsubsection{\textbf{Implementation Details}}
We implemented our proposed model and all baselines using PyTorch on an Ubuntu 24.04 LTS system equipped with an NVIDIA GeForce RTX 3060 GPU. All input features were normalized using min-max scaling. During training, we minimized the loss function using the Adam optimizer \cite{kingma2014adam} with an initial learning rate of $\alpha =0.0001, \beta_{1}^{} =0.9, \beta_{2}^{} =0.999$. The batch size was set to 32, and the dynamic weight averaging parameter was 2. For data augmentation, the perturbation rates for traffic data and graph topology were set to 0.2 and 0.5, respectively. The hidden layer dimension for eKamba was set to 512, and the GCN convolution kernel was configured to 3x3. To prevent overfitting, we employed an early stopping technique.

\subsubsection{\textbf{Baselines and SSL-eKamba variants}}
We compare our proposed model with the following existing baselines for traffic accident prediction. For each method, we carefully tuned the key parameters to ensure optimal performance. To highlight the contributions of our approach, we also designed a generic baseline model called SSL-eKamba-Base.

\begin{itemize}
    \item 
SSL-eKamba-Base: excludes the following two components compared to SSL-eKamba, w/o eKAN: Replaces eKAN with a fully connected (FC) layer.
w/o Self-Supervised Task: Omits the self-supervised task, ignoring spatio-temporal heterogeneity.
\item XGBoost \cite{chen2016xgboost}: A gradient boosting framework for regression, with 100 estimators and a max depth of 7.
\item ConvLSTM \cite{shi2015convolutional}: Combines CNNs and LSTMs for spatio-temporal prediction, with a 3x3 CNN kernel, 64 filters, and 64 LSTM hidden units.
\item  GSNet \cite{wang2021gsnet}: A deep learning model that combines Graph Neural Networks (GCN) and GRU, designed for tasks with complex spatio-temporal dependencies. The GCN kernel size is set to 3x3, GCN filters to 64, and GRU hidden layers to 256.
\item SARM \cite{saleh2022traffic}: A traffic accident predictor based on the transformer architecture, introducing multi-view processing for spatio-temporal variability. Attention Heads are set to 8, and Position Embedding is set to 64.
\item CFC \cite{grigorev2023traffic}: An extended version of SARM, incorporating a multi-scale transformer architecture for fine-grained accident risk prediction.
\item MVMT-STN \cite{wang2021traffic}: An advanced multi-task deep learning framework for traffic accident risk prediction, based on multi-granularity GCN.
\item RiskSeq \cite{zhou2020foresee}: Combines differential time-varying GCN with a context-guided LSTM decoder for fine-grained traffic prediction. GCN kernel: 3x3, filters: 256.
\item iTransformer \cite{liu2023itransformer}: Analyzes individual variable time series first, then merges them, making it SOTA in time series forecasting.
\item STGCN \cite{song2020spatial}: A multi-step traffic prediction model using stacked spatio-temporal convolutional blocks. Each block has three layers with 64 filters.
\end{itemize}
\begin{table}[]
\setlength{\tabcolsep}{6.8mm}
\caption{PERFORMANCE COMPARISON OF DIFFERENT APPROACHES}
    \label{table2}
\begin{tabular}{@{}l|ccc@{}}
\toprule
\multicolumn{4}{c}{\textbf{Dataset}} \\ 
\multicolumn{4}{c}{\textit{(a) New York City}}  \\\midrule
{Method}       & RMSE   & Recall  & MAP    \\ \midrule
{XGBoost}           & 11.2655 & 23.31\% & 0.1075\\
{ConvLSTM}    & 8.1502 & 29.81\% & 0.1426 \\
{STGCN}        & 7.6212 & 33.29\% & 0.1753 \\
{GSNet}        & 7.6649 & 33.58\% & 0.1785 \\
{RiskSeq}     & 7.3423 & 34.52\% & 0.1887 \\
{iTransformer}& 7.7831 & 32.18\% & 0.1696 \\
{SARM}        & 7.2138 & 34.82\% & 0.1926 \\
{CFC}        & 6.9746 & 33.69\% & 0.1836 \\
{MVMT-STN}    & 7.3132 & 34.96\% & 0.1915 \\ \midrule
{SSL-eKamba-Base}           & 7.9074 & 31.59\% & 0.1582 \\
{\textbf{SSL-eKamba}} & \textbf{6.6839} & \textbf{35.23\%} & \textbf{0.1948} \\ \midrule
\multicolumn{4}{c}{\textit{(b) Chicago}}            \\ \midrule
{XGBoost}           & 15.7835 & 12.36\% & 0.0517 \\
{ConvLSTM}    & 11.4573 & 18.19\% & 0.0716 \\
{STGCN}        & 10.2155 & 19.56\% & 0.0918 \\
{GSNet}        & 10.7694 & 20.14\% & 0.0917 \\
{RiskSeq}     & 10.2734 & 20.52\% & 0.0905 \\
{iTransformer}& 9.7326 & 21.42\% & 0.1075 \\
{SARM}        & 9.3857 & 22.37\% & 0.1095 \\
{CFC}        & 9.1513 & 22.94\% & 0.1053 \\
{MVMT-STN}    & 9.6824 & 21.77\% & 0.0928 \\ \midrule
{SSL-eKamba-Base}           & 11.0187 & 17.59\% & 0.0883 \\
{\textbf{SSL-eKamba}} & \textbf{9.0678} & \textbf{22.61\%} & \textbf{0.1134} \\ \bottomrule

\end{tabular}
\end{table}

\subsection{Result analysis}
\label{sec_e_B}
Table \ref{table2} and Figures xx, xx compare the performance of our SSL-eKamba model with various baseline models. Traffic accident prediction models encounter several challenges in real-world deployment, such as generalizability across different cities, temporal heterogeneity within the same city, and the need for real-time processing. Shallow models are efficient but often lack accuracy, while more complex models improve accuracy but struggle to meet real-time demands.

SSL-eKamba successfully balances these trade-offs by employing a self-supervised learning paradigm and a novel model architecture, achieving superior or comparable performance across different tasks. The specific results for each experiments are summarized below:

\subsubsection{\textbf{Traffic Accident Risk Prediction Analysis}}
In terms of predictive accuracy, SSL-eKamba achieved the best results across all datasets, highlighting its effectiveness in jointly modeling spatial and temporal heterogeneity in a self-supervised manner. From table \ref{table2}, we can draw the following conclusions:

For traditional models like XGBoost (XGB) and LSTM, their inherent structural simplicity offers an advantage in computational efficiency. However, these models struggle to adequately handle large-scale complex data, particularly high-dimensional data, due to their limited nonlinear representational capabilities. Consequently, their accuracy is lower compared to more advanced methods, such as iTransformer.
However, all three models—XGBoost, ConvLSTM and iTransformer—only consider temporal dependencies, which likely contributes to their average MAP score of 14.71\%, 4.78\% lower than our model. In contrast to these methods, GSNet,\begin{figure}[htbp]
	\centering
        
	\begin{minipage}[c]{0.24\textwidth}
		\centering
		\includegraphics[width=\textwidth]{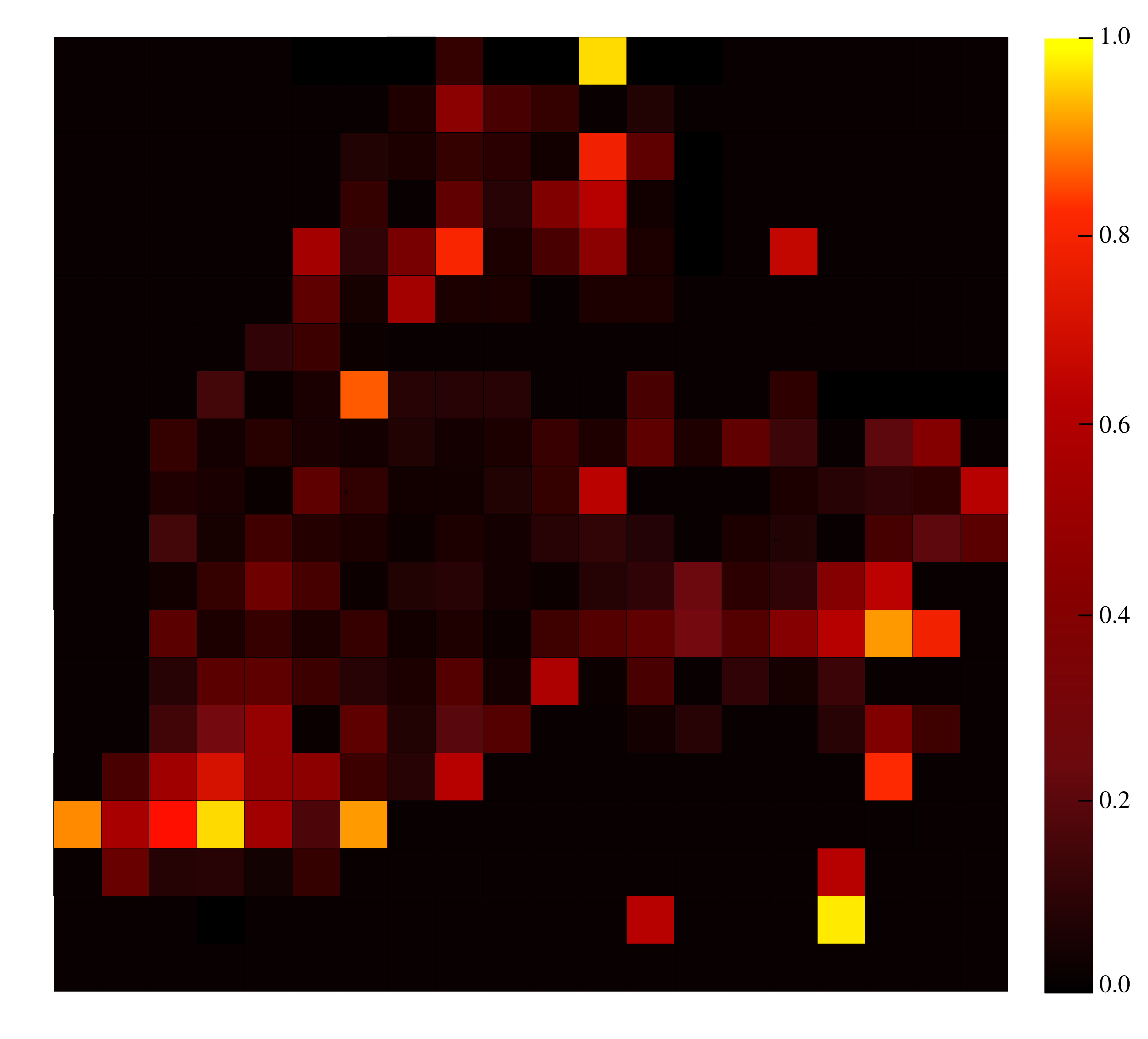}
		\caption*{(a) SSL-eKamba-base}
		\label{fig_ef}
	\end{minipage} 
	\begin{minipage}[c]{0.24\textwidth}
		\centering
		\includegraphics[width=\textwidth]{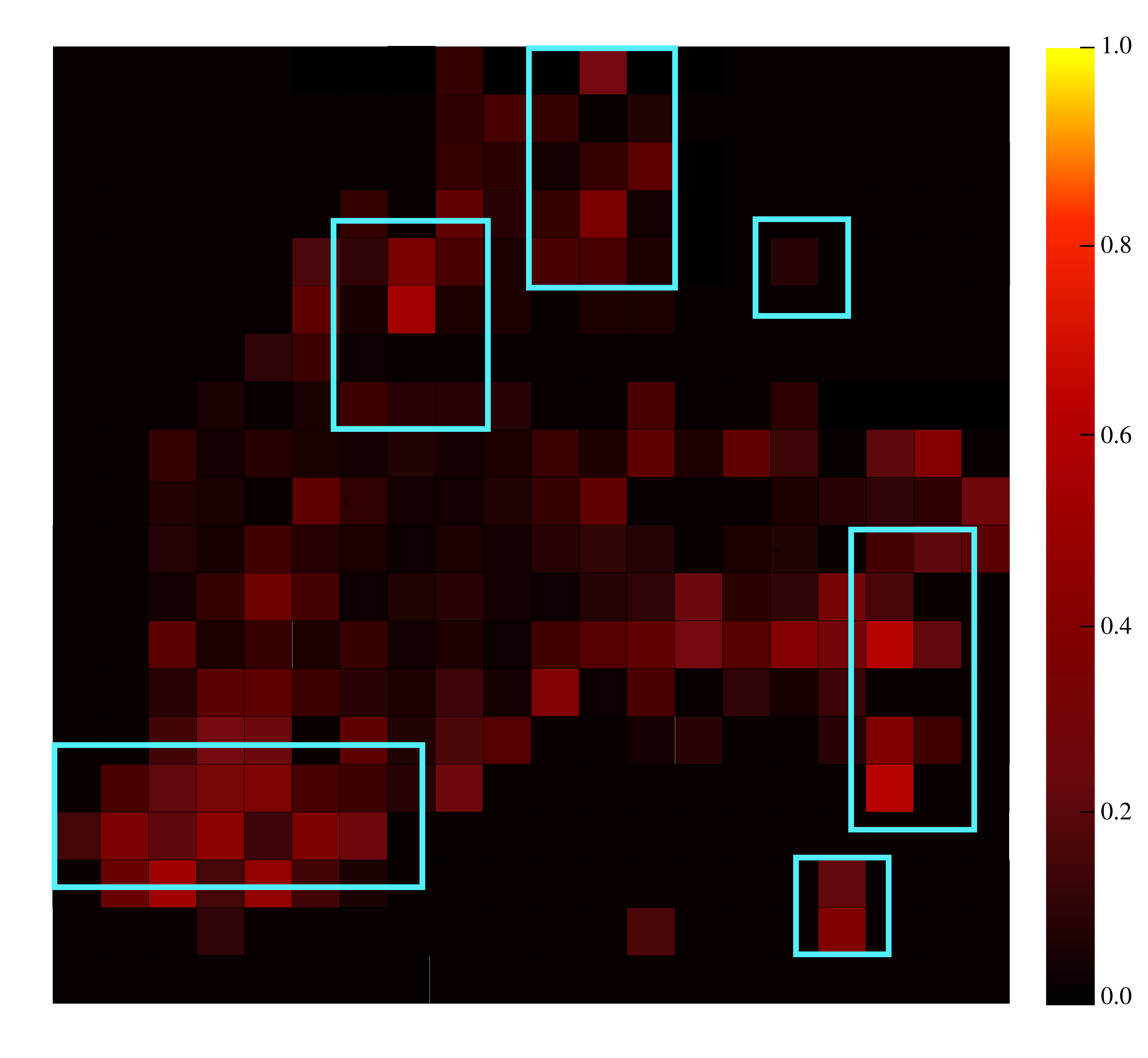}
		\caption*{(b) SSL-eKamba}
		\label{(b) SSL-eKamba}
	\end{minipage}
        \centering
	\begin{minipage}[c]{0.24\textwidth}
		\centering
		\includegraphics[width=\textwidth]{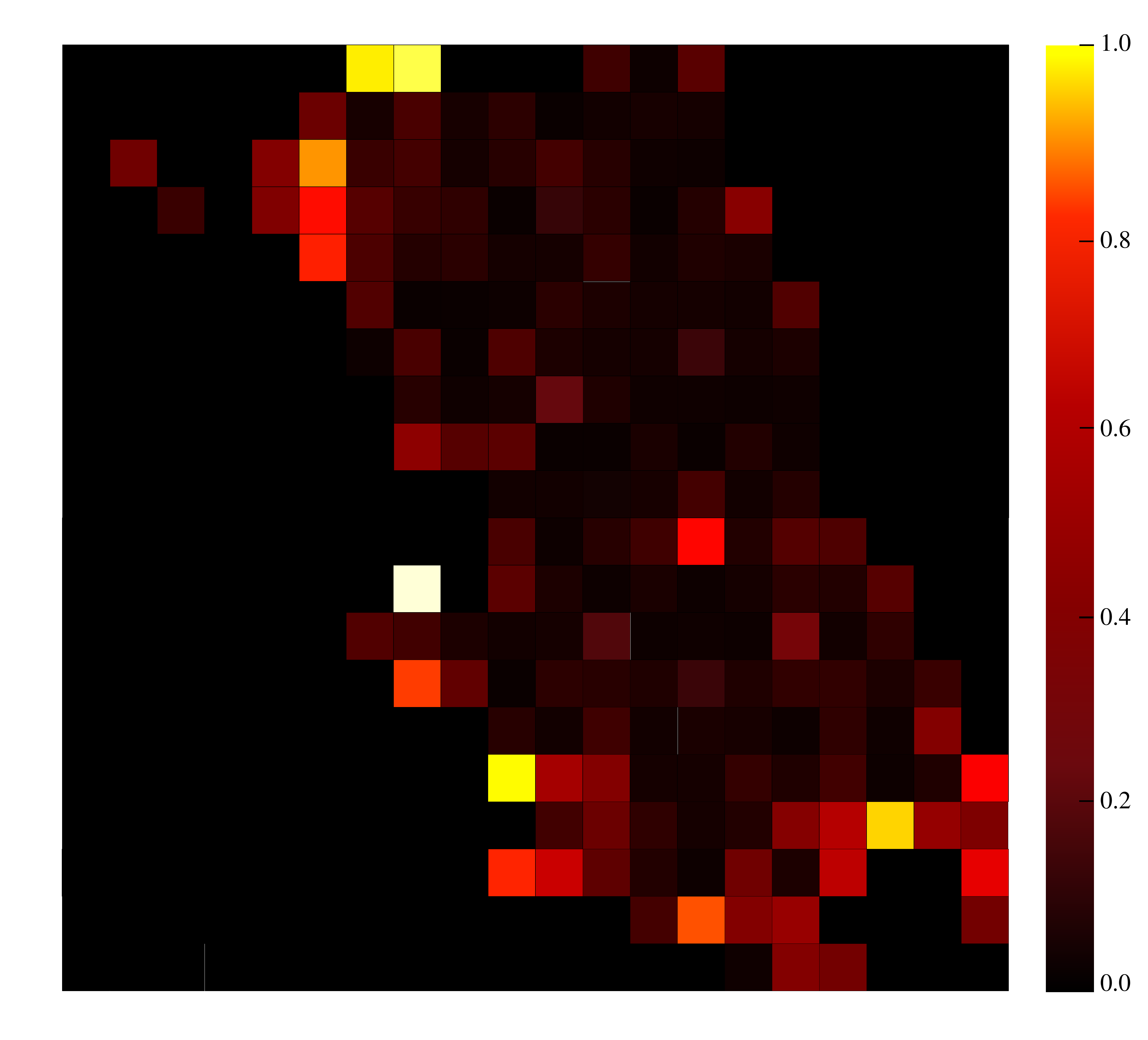}
		\caption*{(c) SSL-eKamba-base}
		\label{fig_E3_1}
	\end{minipage} 
	\begin{minipage}[c]{0.24\textwidth}
		\centering
		\includegraphics[width=\textwidth]{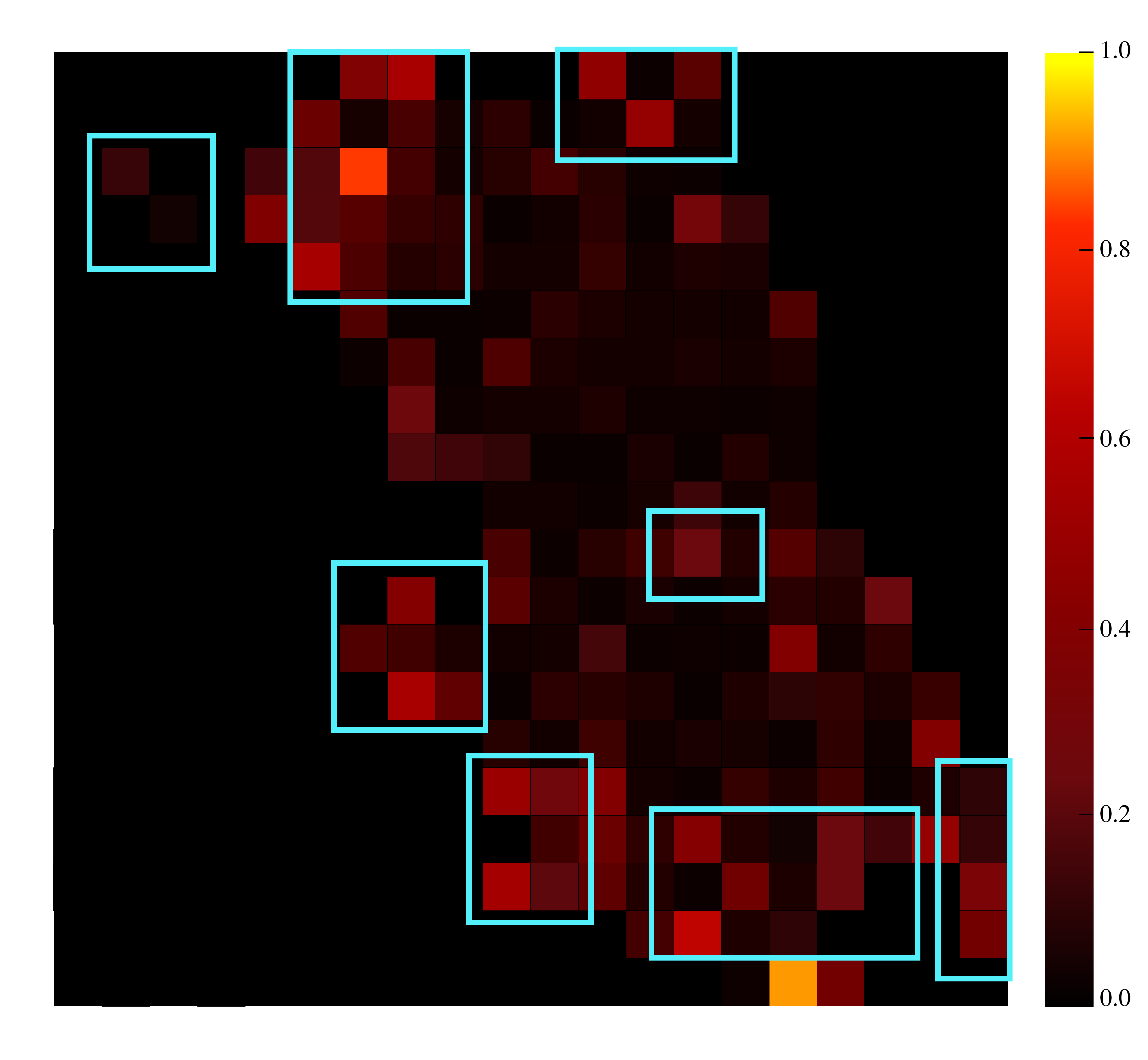}
		\caption*{(d) SSL-eKamba-base}
		\label{fig_E3_2}
	\end{minipage}
 \caption{Visualizing traffic accident prediction error results separately for the two datasets.}\label{fig4}
\end{figure} STGCN, and SARM account for both temporal and spatial dependencies, resulting in better MAP scores. However, these models still struggle to fully capture complex spatio-temporal dependencies due to their reliance on single or static learning approaches.

RiskSeq, MVMT-STN, and CFC go further by exploring more advanced techniques to capture spatio-temporal dependencies efficiently. Yet, their multi-scale, multi-view learning approaches depend heavily on manually collected features, which may overlook implicit patterns or correlations within the data. This reliance can limit the models' generalizability across different cities.

Unlike the above methods, our approach utilizes self-supervised learning to adaptively capture spatio-temporal dependencies directly from traffic data.Fig \ref{fig4} presents the prediction errors and visualization results across different datasets, where brighter pixels indicate larger errors. Consistent with the quantitative results in Table \ref{table2}, our model demonstrates consistent performance improvements, underscoring the effectiveness of SSL-eKamba across various urban environments.

Moreover, we observed significant improvements in boxed areas within Figure 1, which correspond to regions with sparser data. This unexpected result suggests that our self-supervised, spatio-temporal adaptive modeling is particularly effective in transmitting information across similar areas within a city, further enhancing its applicability in diverse urban environments.

\subsubsection{\textbf{Generalizability Analysis}}
To further investigate the generalizability of SSL-eKamba, we selected CFC and MVMT-STN,  the most recent state-of-the-art methods in traffic accident prediction, using the NYC dataset. These methods\begin{figure}[htbp]
	\centering
        
	\begin{minipage}[c]{0.24\textwidth}
		\centering
		\includegraphics[width=0.9\textwidth,height=3.3cm]{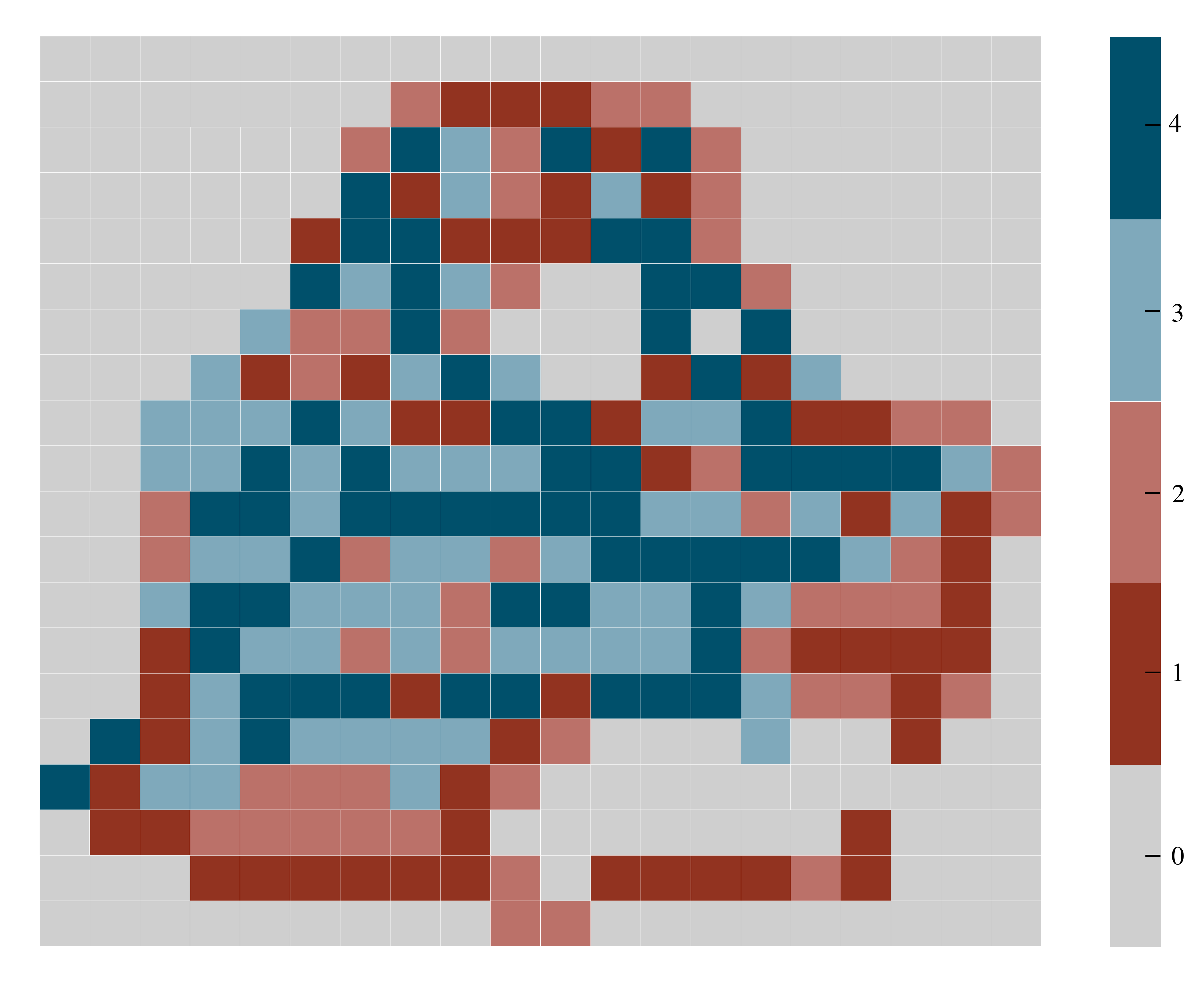}
		\caption*{(a) Spatial Classify}
		\label{fig_ef}
	\end{minipage} 
	\begin{minipage}[c]{0.24\textwidth}
		\centering
		 \resizebox{1\textwidth}{!}{ 
        	\begin{tabular}{@{}ccc@{}}
\toprule
\textbf{Type}                                  & \textbf{Period} & \textbf{Classify} \\ \midrule
\multicolumn{1}{c|}{\multirow{4}{*}{Workday}}  & 7:00-10:00      & 0                 \\
\multicolumn{1}{c|}{}                          & 10:00-17:00     & 1                 \\
\multicolumn{1}{c|}{}                          & 17:00-20:00     & 2                 \\
\multicolumn{1}{c|}{}                          & 20:00-7:00      & 3                 \\ \midrule
\multicolumn{1}{c|}{\multirow{2}{*}{Vacation}} & 9:00-22:00      & 4                 \\
\multicolumn{1}{c|}{}                          & 22:00-9:00      & 5                 \\ \bottomrule
\end{tabular}}
            \captionsetup{skip=18pt} 
		\caption*{(b) Temporal Classify}
		\label{(b) SSL-eKamba}
	\end{minipage}
        \centering
	\begin{minipage}[c]{0.24\textwidth}
		\centering
		\includegraphics[width=\textwidth]{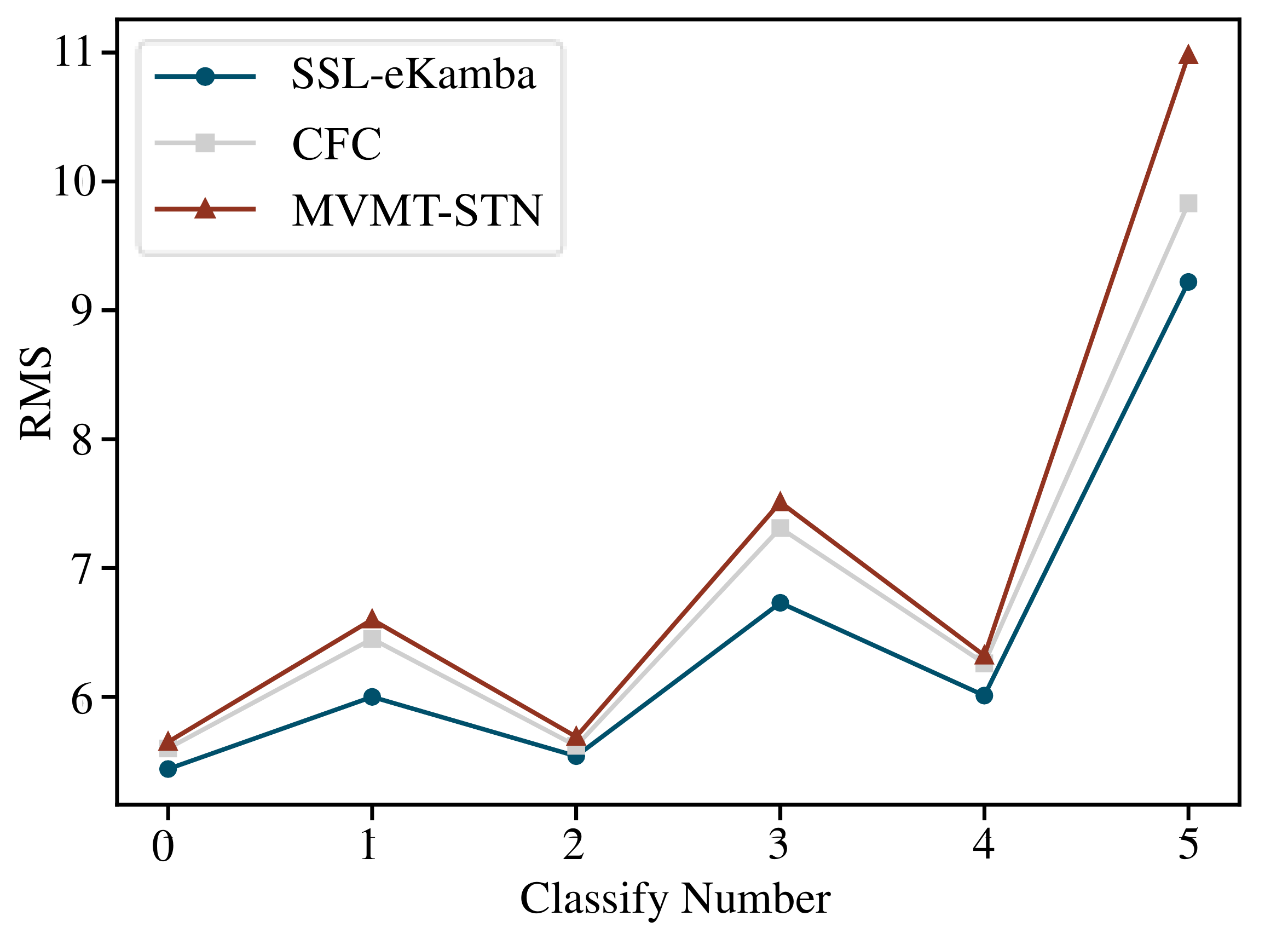}
		\caption*{(c) Spatial Performance }
		\label{fig_E3_1}
	\end{minipage} 
	\begin{minipage}[c]{0.24\textwidth}
		\centering
		\includegraphics[width=\textwidth]{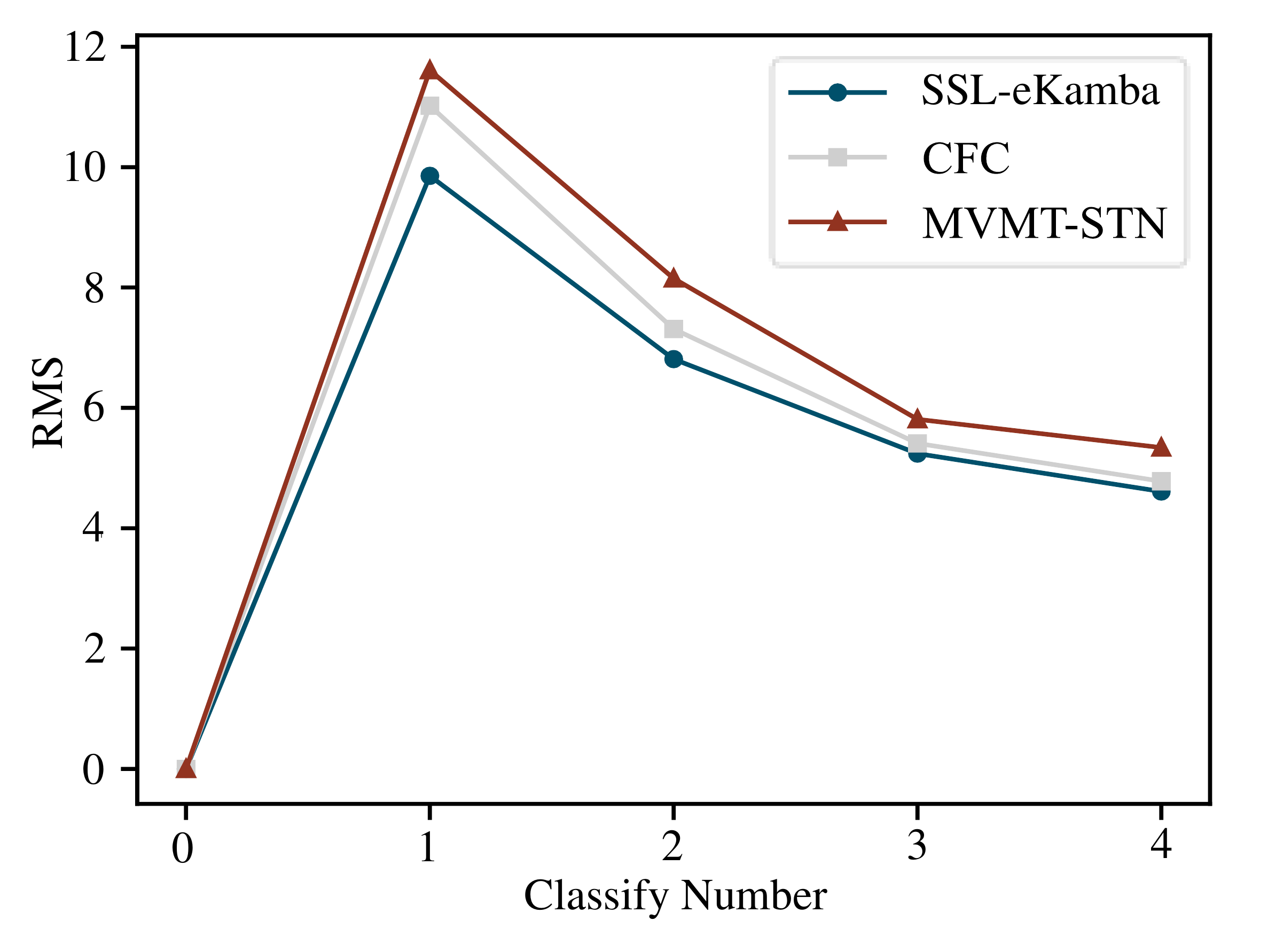}
		\caption*{(d) Temporal Performance}
		\label{fig_E3_2}
	\end{minipage}
 \caption{Model performance across different time periods and regions}\label{fig5}
\end{figure}
were employed to predict traffic accidents in spatial regions with heterogeneous data distributions and across time periods with varying patterns. Specifically, we clustered regions using traffic data statistics, such as mean, median, and standard deviation.

As depicted in Fig \ref{fig5} (a), regions with lower id values correspond to areas with sparser data, which typically have lower accident rates. Fig \ref{fig5} (b) shows the prediction performance for different classify. Consistent with the results in Fig \ref{fig4}, our SSL-eKamba model consistently outperforms the baselines, particularly in regions with fewer accidents. For example, in classify 1 and 2, the performance improves by an average of 12.94\% and 11.9\%, respectively, compared to the baseline models.
We attribute the baseline model's limitations to its reliance on manual feature engineering, which may overlook implicit patterns or associations within the data, particularly when performing multi-scale and multi-view learning. These subtle patterns may not be intuitively visible or easily understood.

For temporal heterogeneity, we divided the traffic rhythm into six time segments based on previous studies [Wang, GSNet], as shown in Fig \ref{fig5} (c). Weekdays were divided into four time segments, while holidays were divided into two segments. Fig \ref{fig5} (d) shows the prediction performance across these time periods. Our SSL-eKamba model consistently outperforms the baselines across all segments, with significant improvements during non-peak traffic periods—specifically in time periods 1, 3, and 5—where the RMSE improved by an average of 8.75\%, 10.1\%, and 12.85\%, respectively. These performance gains are likely due to the data enhancement techniques employed and the self-supervised learning paradigm.
(For further analysis, please refer to Section xx, Ablation Studies.)
This further validates the robustness of the SSL-eKamba model in predicting traffic accidents across diverse temporal and spatial regions. The model's ability to maintain predictive accuracy and stability, even under varying environmental conditions and potential anomalies, highlights its effectiveness in real-world scenarios.

\begin{table}[]
\setlength{\tabcolsep}{3.3mm}
\caption{ABLATION RESULTS FOR KEY CONTRIBUTORS IN THE NYC DATASET}
    \label{table3}
\begin{tabular}{@{}ll|cccl@{}}
\toprule
\multicolumn{2}{c|}{\textbf{Components}} &        &                              &        &  \\
\multicolumn{1}{l|}{eKAN} & \multicolumn{1}{c|}{Self-Supervised} & \multirow{-2}{*}{RMSE} & \multirow{-2}{*}{Recall} & \multirow{-2}{*}{MAP} &  \\ \midrule
\multicolumn{1}{c}{\ding{55}} & \multicolumn{1}{c|}{\ding{55}}  & 7.9074 & 31.59\% & 0.1582 & $\diamondsuit$ \\
\multicolumn{1}{c}{\ding{55}} & \multicolumn{1}{c|}{\ding{51}} & 7.3849 & 34.66\% & 0.1907 &  \\
\multicolumn{1}{c}{\ding{51}} & \multicolumn{1}{c|}{\ding{55}} & 7.6212 & 33.21\% & 0.1753 &  \\
\multicolumn{1}{c}{\ding{51}} & \multicolumn{1}{c|}{\ding{51}} & \textbf{6.6839}  & \textbf{35.23}\% & \textbf{0.1948} & $\heartsuit$ \\ \bottomrule
\end{tabular}

\begin{footnotesize}
$\diamondsuit$: SSL-eKamba-base $\heartsuit$: SSL-eKamba
\end{footnotesize}
\end{table}

\begin{table}[]
\setlength{\tabcolsep}{7.1mm}
\caption{DETAILED ABLATION OF THE SELF-SUPERVISED AUXILIARY TRAINING TASKS}
    \label{table4}
\begin{tabular}{@{}c|ccc@{}}
\toprule
\textbf{Components} & \multirow{2}{*}{RMSE} & \multirow{2}{*}{Recall} & \multirow{2}{*}{MAP} \\
Self-Supervised   &       &         &        \\ \midrule
w/o SSL-Spatial  & 7.518 & 33.67\% & 0.1798 \\
w/o SSL-Temperal & 7.462 & 33.92\% & 0.1834 \\
SSL-eKamba       & 6.683 & 35.23\% & 0.1948 \\ \bottomrule
\end{tabular}
\end{table}

\subsection{Ablation Study}
\label{sec_e_C}
Given the richer feature set and broader temporal coverage of the NYC dataset, we chose this dataset for the ablation experiments.

\textbf{Primary Ablation Study. }As shown in Table \ref{table3}, we evaluated the contribution of each component in SSL-eKamba by sequentially removing them from the model. The results indicate that utilizing the spatio-temporal self-supervised differential learning module to assist in training eKamba is effective. Notably, when both components are enabled, the RMSE significantly decreases to 6.683, representing the highest performance improvement. This suggests that the two components complement each other during model training. Table \ref{table4} presents the results for the model variants without temporal heterogeneity (w/o Temporal) and spatial heterogeneity (w/o Spatial) in the self-supervised tasks. SSL-eKamba consistently outperforms both variants, demonstrating the necessity of jointly modeling spatial and temporal heterogeneity.

\textbf{Why Is the Improvement More Noticeable in Sparse Data Regions? } While the previous experiments confirmed the effectiveness of the SSL-eKamba model across different cities, the visualized results in Fig \ref{fig4} (b) and Fig \ref{fig4} (d) unexpectedly revealed that the most significant performance improvements stem from a marked reduction in prediction errors in data-sparse regions.

To further evaluate the impact on SSL-eKamba, we conducted ablation experiments by replacing the heterogeneity-guided data augmentation with random edge and weight enhancements. As shown in Table \ref{table5}, the performance of both experimental groups with the modified data augmentation was inferior to that of Kamba (w/o SSL) and SSL-eKamba. This indicates the effectiveness of our designed adaptive data augmentation, whereas random data introduces misleading information, such as incorrect regional adjacency relationships, leading to increased prediction errors.
Moreover, in the second experimental group, where the self-supervised tasks were retained, the performance remained better than in the first group, with RMSE values of 7.846 and 7.357 compared to 8.923 and 8.301. This suggests that self-supervised learning can still enhance the model’s generalization capability, even in the presence of noisy data.

From these findings, we can draw the following conclusions:
a) The two adaptive data augmentation techniques designed for the self-supervised auxiliary training tasks indirectly provide the model with more effective data samples for training.
b) Self-supervised auxiliary training not only effectively models spatio-temporal dependencies across different cities but also captures spatio-temporal variability within various regions of an entire city.

\begin{table}[]
\setlength{\tabcolsep}{6mm}
\caption{DETAILED ABLATION OF OUR ADAPTIVE DATA AUGMENTATION}
    \label{table5}
\begin{tabular}{@{}cccc@{}}
\toprule
\multicolumn{4}{c}{\textit{(a) w/o SSL-Supervised}  }                        \\ \midrule
\multicolumn{1}{c|}{\textbf{Components}} & \multirow{2}{*}{RMSE} & \multirow{2}{*}{Recall} & \multirow{2}{*}{MAP} \\
\multicolumn{1}{c|}{Data Augmentation} &        &         &        \\ \midrule
\multicolumn{1}{c|}{Random Edge}       & 8.4223 & 27.54\% & 0.1137 \\
\multicolumn{1}{c|}{Random Weigth}     & 7.9359 & 30.42\% & 0.1494 \\
\multicolumn{1}{c|}{eKamba}            & 7.6212 & 33.21\% & 0.1753 \\ \midrule
\multicolumn{4}{c}{\textit{(b) SSL-Supervised}   }                           \\ \midrule
\multicolumn{1}{c|}{Random Edge}       & 7.846  & 32.78\% & 0.1604 \\
\multicolumn{1}{c|}{Random Weigth}     & 7.357  & 33.47\% & 0.1799 \\
\multicolumn{1}{c|}{SSL-eKamba}        & 6.683  & 35.23\% & 0.1948 \\ \bottomrule
\end{tabular}
\end{table}

\subsection{Model Efficiency Analysis}
\label{sec_e_D}
\begin{figure*}[htbp]
	\centering
	\begin{minipage}{0.49\linewidth}
		\centering
		\includegraphics[width=\textwidth]{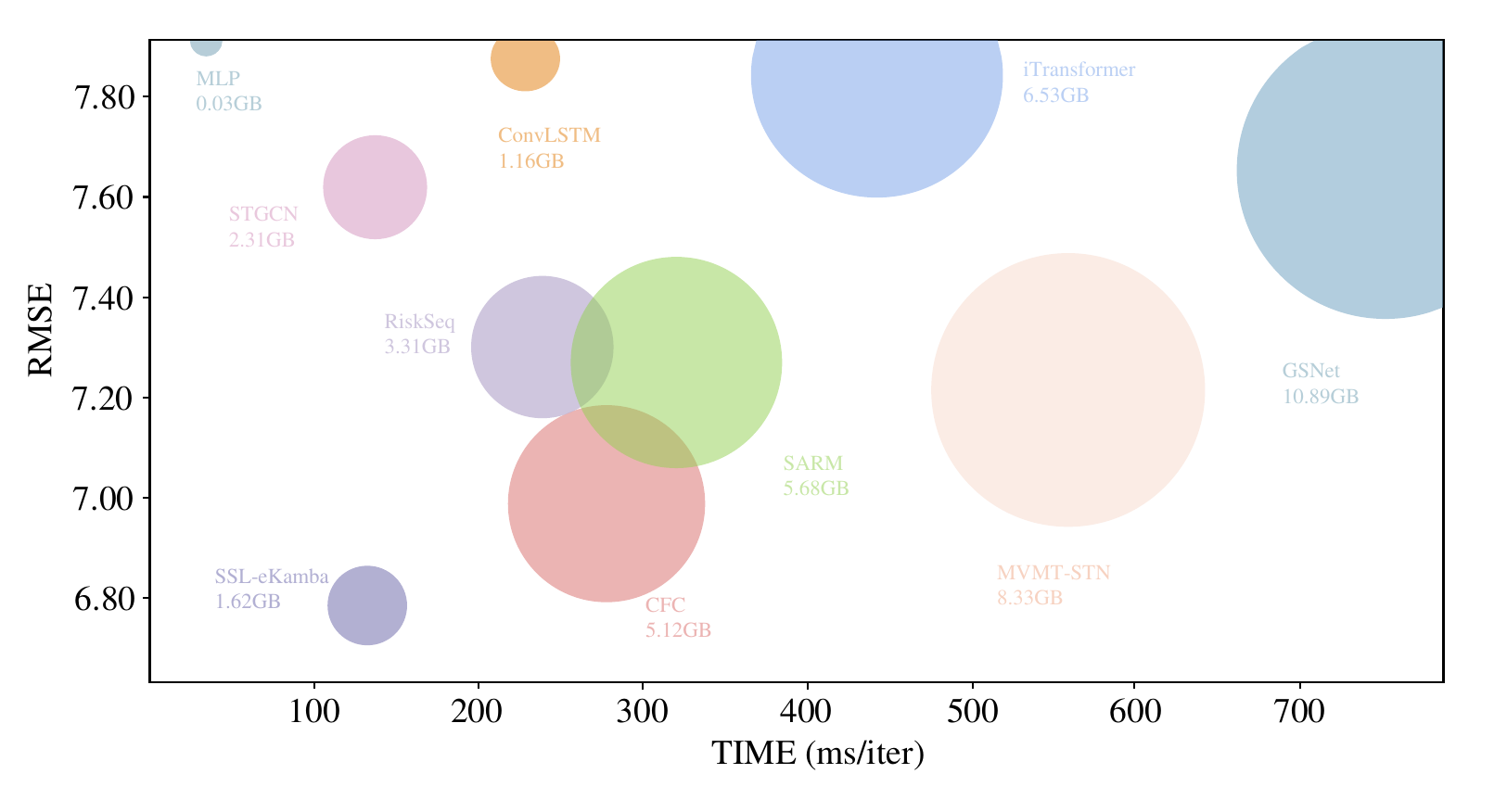}
		\caption*{(a) New York City}
		\label{fig_ef}
		\label{chutian1}
	\end{minipage}
	\begin{minipage}{0.49\linewidth}
		\centering
		\vspace{-1pt}
		\includegraphics[width=\textwidth]{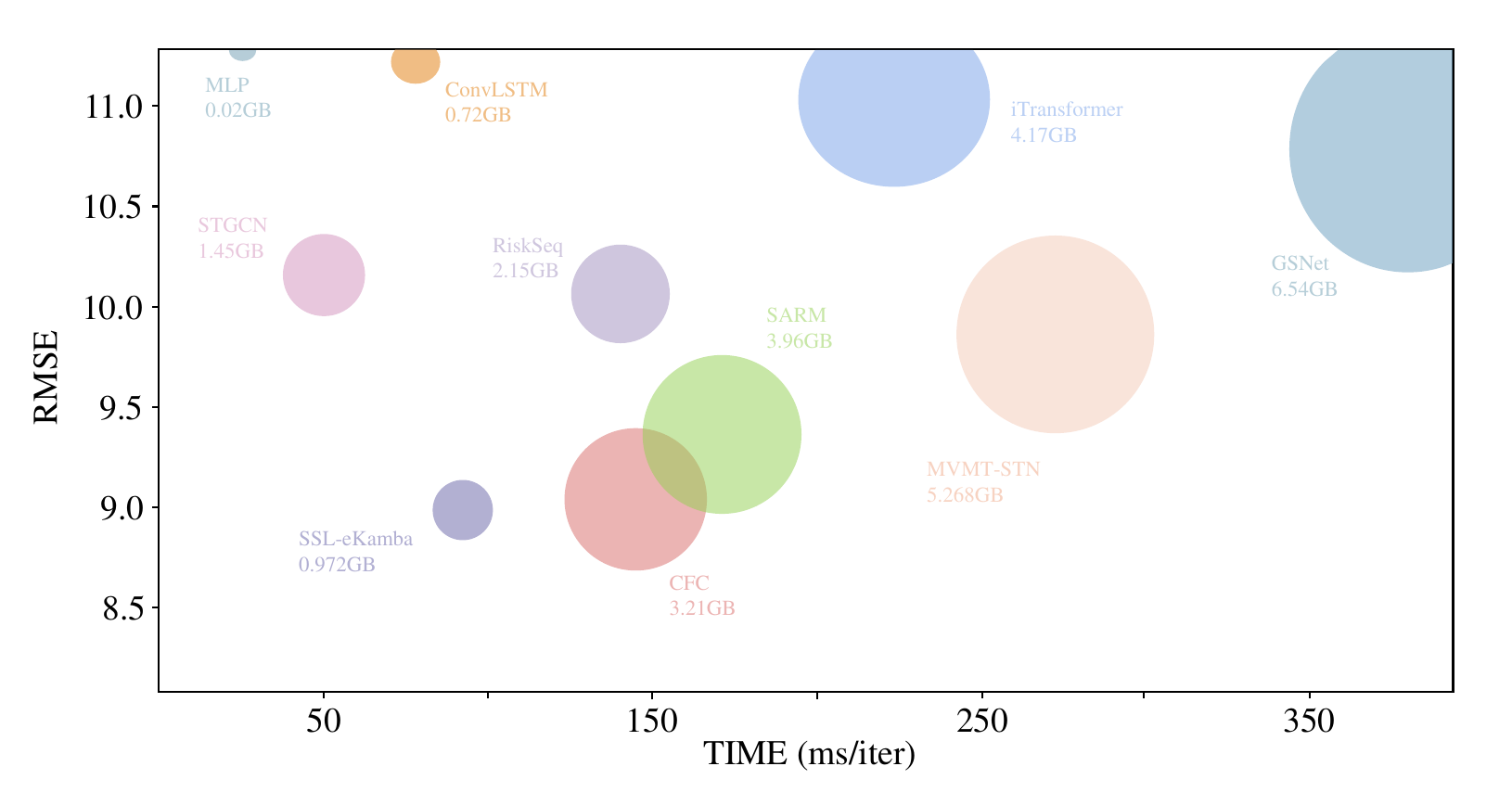}
        \captionsetup{skip=10pt} 
		\caption*{(b) Chicago}
		\label{(b) SSL-eKamba}
	\end{minipage}
 \caption{Comparison of SSL-eKamba and baselines on RMSE, TIME, and GPU Memory.}\label{fig6}
\end{figure*}

\textbf{Analysis of efficiency results. }To evaluate the computational efficiency of SSL-eKamba, we compared its memory usage and computation time against several baseline models on the NYC and Chicago datasets. Independent runs were conducted on a single NVIDIA RTX 3060 GPU, with a batch size set to 32. The detailed results are presented in Fig  \ref{fig6}.

In our study, the vertical axis represents the Root Mean Square Error (RMSE), while the horizontal axis quantifies the model's computation time (average cumulative wall-clock time per epoch for all samples with the same batch size). The size of the bubbles corresponds to the GPU memory allocation.
SSL-eKamba achieved the most favorable RMSE metrics on both the NYC and Chicago datasets. While the original deep learning models, such as MLP and LSTM, demonstrated significant advantages in computation speed and memory usage, their RMSE performance was limited. This limitation likely arises from the complexity of traffic datasets, which typically contain intricate nonlinear patterns and interactions that shallow models struggle to capture effectively.

Surprisingly, the recently proposed advanced model iTransformer also exhibited limited performance on this task. We hypothesize that general-purpose models, not specifically tailored to the nuances of the task, often struggle to achieve superior results.
STGCN, on the other hand, benefits from its convolutional architecture, which reduces the number of parameters and supports parallel training, leading to better efficiency compared to other baseline models. However, it still lacks a competitive edge in predictive accuracy.
Models based on recurrent neural network (RNN) architectures, such as RiskSeq, MVMT-STN, and GSNet, have been specifically adapted for the traffic accident prediction task. While these adaptations have improved predictive performance, the RNN structure's inherent sequential data processing limits its ability to handle parallel tasks, resulting in lower training efficiency. This inefficiency is particularly pronounced in MVMT-STN and GSNet, where the introduction of attention mechanisms, although effective, further increases computational demands due to the need to explicitly store the entire context, leading to significant resource consumption during both training and inference.
Similarly, models like SARM and CFC, which are based on transformer architectures, also suffer from the limitations of attention mechanisms. These models require considerable memory and computational resources during operation.

In contrast, SSL-eKamba generally requires shorter inference times and less GPU memory allocation compared to the baseline models, making it more suitable for real-time applications.

\textbf{Why Does SSL-eKamba Achieve a Good Balance Between Performance and Efficiency? }To further investigate the contributions of the model, we analyzed the NYC dataset by replacing the linear layers in Mamba. The original Mamba structure with linear layers is indicated in red. As shown in Fig \ref{fig6}, SSL-eKamba significantly reduces computational overhead compared to other baselines while maintaining superior performance.

\begin{figure}[htbp]
	\centering
	\begin{minipage}{0.7\linewidth}
		\centering
		\includegraphics[width=\textwidth]{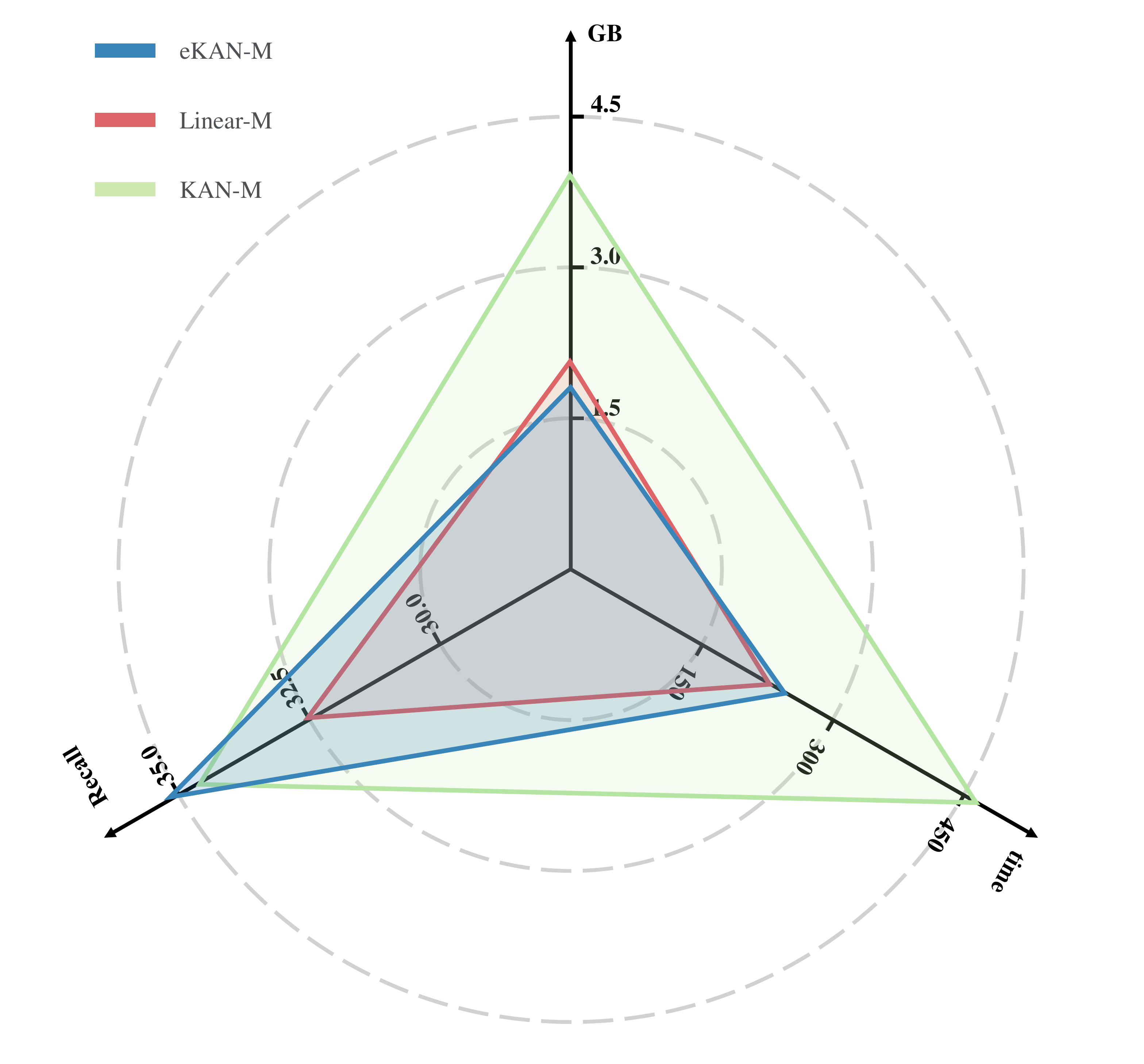}
		
	\end{minipage}
 \caption{Investigating the impact of different encoding layers on the model's nonlinear representation capability.}\label{fig7}
 
\end{figure}

We attribute this balance partly to the Mamba architecture, which incorporates a SelectiveSSM mechanism. This mechanism allows the model to selectively process input information by parameterizing the inputs to the SSM, enabling the model to focus on or ignore specific inputs. This is in contrast to the attention mechanism in transformers, which requires processing the entire context, leading to higher computational costs.
However, we recognize that this selective mechanism may reduce the model's capacity. As shown in Fig \ref{fig7}, the experiment demonstrates that the recall score (red) is only 32.52\%, lower than other models. When we introduced the KAN network, which has stronger modeling capabilities, to replace the linear layers, the performance improved, with the recall score (green) rising to 35.37\% in Fig \ref{fig7}. This experiment suggests that enhancing the modeling capacity of the linear layers can partially offset the negative impact of Selective SSM.

However, the original KAN network is not well-suited for real-time tasks like traffic accident prediction. The green bars in Fig \ref{fig7} show that both GPU memory usage (GB) and computation time (time) more than doubled. We attribute this efficiency issue to the need for KAN to expand all intermediate variables to execute different activation functions (e.g., B-spline functions).

In contrast, the blue bars in Fig \ref{fig7} demonstrate that eKAN-Mamba maintains the recall performance while significantly improving both GB and time compared to KAN-Mamba. This indicates that our improvements to KAN not only resulted in substantial gains in computational efficiency but also preserved its modeling capabilities.

In conclusion, by enhancing the modeling capacity of the linear layers in Mamba with eKAN, the model can achieve improved prediction performance while maintaining computational efficiency.

\section{CONCLUSION AND FUTURE WORK}
\label{sec5}
In this work, we addressed previously unexplored challenges in the practical application of traffic accident prediction. This is a significant exploration. To tackle the issue of generalization, we proposed a spatio-temporal self-supervised learning framework that enables the model to adaptively capture spatio-temporal heterogeneity signals, supplementing the primary traffic accident prediction task. For the challenge of real-time performance, we restructured the computational architecture of the state-space model using the Kolmogorov-Arnold framework to encode information across space and time, thereby improving computational efficiency. The proposed model consistently outperformed state-of-the-art methods on two real-world datasets, demonstrating the robustness of SSL-eKamba.

For future work, we plan to refine our model to the road-level for more precise accident location prediction. Additionally, we will explore integrating a decomposed spatio-temporal heterogeneity self-supervised framework to build a cohesive spatio-temporal collaborative self-supervised model.


\bibliographystyle{unsrt}
\bibliography{hyperref}

\end{document}